\documentclass[10pt,twocolumn,a4paper]{article}

\usepackage[final,pagenumbers]{cvww}

\usepackage{makecell}
\usepackage{subcaption}
\usepackage[export]{adjustbox}

\usepackage{graphicx}
\usepackage{amsmath}
\usepackage{amssymb}
\usepackage{booktabs}
\usepackage{enumitem}
\usepackage{fancyhdr}

\usepackage{amsthm}
\usepackage{pifont}% http://ctan.org/pkg/pifont
\usepackage[dvipsnames]{xcolor}
\usepackage{colortbl}
\usepackage{pgfplots}
\pgfplotsset{compat=1.18}
\usepackage{subcaption}
\usepackage{multirow}
\usepackage{lipsum}

\usepackage[accsupp]{axessibility}  % Improves PDF readability for those with disabilities.

\definecolor{cvprblue}{rgb}{0.21,0.49,0.74}
\usepackage[pagebackref,breaklinks,colorlinks,allcolors=cvwwblue]{hyperref}

\usepackage[capitalize]{cleveref}
\crefname{section}{Sec.}{Secs.}
\Crefname{section}{Section}{Sections}
\Crefname{table}{Table}{Tables}
\crefname{table}{Tab.}{Tabs.}

\newcommand{\cmark}{\ding{51}}%
\newcommand{\xmark}{\ding{55}}%

\definecolor{LightCyan}{rgb}{0.7,1,1}
\definecolor{LightYellow}{rgb}{1,1,0.7}
\definecolor{LightOrange}{rgb}{1,0.8,0.5}
\definecolor{DarkGreen}{rgb}{0.0,0.5,0.0}
\definecolor{Burgundy}{rgb}{0.5, 0, 0.125}
\definecolor{HunterGreen}{rgb}{0.207, 0.367, 0.23}
\definecolor{PastelGreen}{rgb}{0.9, 1.0, 0.9} % 0.757, 1.0, 0.757 -- original PastelGreen
\definecolor{LightBlue}{RGB}{220, 245, 255}
\definecolor{DarkGray}{RGB}{120, 120, 120}
\definecolor{DarkerGray}{RGB}{60, 60, 60}
\definecolor{BckPlayerBlue}{RGB}{0, 180, 255}
\definecolor{ForePlayerMagenta}{RGB}{255, 0, 255}
\definecolor{BrightBlue}{RGB}{0, 150, 255}
\definecolor{CobaltBlue}{RGB}{0, 71, 17}
\definecolor{BlueGreen}{RGB}{8, 143, 143}

\newcolumntype{A}{>{\columncolor{LightCyan}}r}
\newcolumntype{B}{>{\columncolor{LightYellow}}r}
\newcolumntype{D}{>{\columncolor{LightOrange}}r}

\def\paperID{**} % *** Enter the Paper ID here

\title{SAM-pose2seg: Pose-Guided Human Instance Segmentation in Crowds}

\author{
Constantin Kolomiiets\hspace{3em}Miroslav Purkrabek\hspace{3em}Jiri Matas
\vspace{1.0em}\\
Visual Recognition Group\\
Department of Cybernetics\\
Faculty of Electrical Engineering\\
Czech Technical University in Prague\\
{\tt\small kolomcon@fel.cvut.cz}
}

\begin{document}
% \maketitle
\twocolumn[{%
\renewcommand\twocolumn[1][]{#1}%
\maketitle
\vspace{-0.8cm}
\begin{center}
    \centering
    \captionsetup{type=figure}
    \begin{subfigure}{0.33\linewidth}
        \centering
        \includegraphics[width=\textwidth]{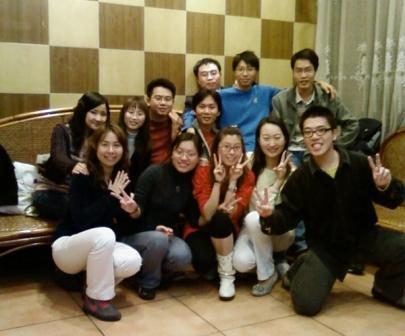}
        \caption*{
        Input image
        }
    \end{subfigure}
    \hfill
    \begin{subfigure}{0.33\linewidth}
        \centering
        \includegraphics[width=\textwidth]{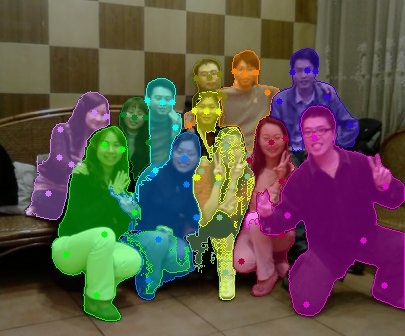}
        \caption*{
        SAM 2.1
        }
    \end{subfigure}
    \hfill
    \begin{subfigure}{0.33\linewidth}
        \centering
        \includegraphics[width=\textwidth]{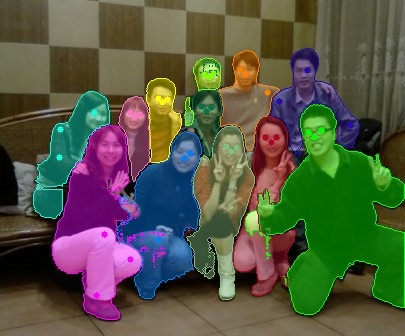}
        \caption*{
        SAM-pose2seg (our)
        }
    \end{subfigure}
    \caption{
    \textbf{SAM-pose2seg is superior to SAM 2} \cite{SAM2} for human instance segmentation, especially in crowded scenes.
    Both examples are generated from the same set of predicted keypoints from \cite{BMPv1}.
    Notice the noise and incorrect masks in the middle of the crowd for SAM 2.1.
    Different prompts are used for SAM 2 and SAM-pose2seg since each model works best with different prompts.
    }
    \label{fig:teaser}
\end{center}%
}]

%%%%%%%%% BODY TEXT
% Best practice is to split the contents of your paper into separate .tex
% files, as shown below:

%%%%%%%%%%%%%%%%%%%%%%%%%%%%%%%%%%%%%%%%%%%% ABSTRACT %%%%%%%%%%%%%%%%%%%%%%%%%%%%%%%%%%%%%%%%
\begin{abstract}
Segment Anything (SAM) provides an unprecedented foundation for human segmentation, but may struggle under occlusion, where keypoints may be partially or fully invisible.
We adapt SAM 2.1 for pose-guided segmentation with minimal encoder modifications, retaining its strong generalization.
Using a fine-tuning strategy called PoseMaskRefine, we incorporate pose keypoints with high visibility into the iterative correction process originally employed by SAM, yielding improved robustness and accuracy across multiple datasets.
During inference, we simplify prompting by selecting only the three keypoints with the highest visibility.
This strategy reduces sensitivity to common errors, such as missing body parts or misclassified clothing, and allows accurate mask prediction from as few as a single keypoint.
Our results demonstrate that pose-guided fine-tuning of SAM enables effective, occlusion-aware human segmentation while preserving the generalization capabilities of the original model.
The code and pretrained models will be available at the \href{https://mirapurkrabek.github.io/BBox-Mask-Pose/}{project website} \footnote{\url{MiraPurkrabek.github.io/BBox-Mask-Pose/}}.
\end{abstract}

%%%%%%%%%%%%%%%%%%%%%%%%%%%%%%%%%%%%%%%%%%%%%%%%%%%%%%%%%%%%%%%%%%%%%%%%%%%%%%%%%%%%%%%%%%%%%%%
%%%%%%%%%%%%%%%%%%%%%%%%%%%%%%%%%%%%%%%%%%%%%%%%%%%%%%%%%%%%%%%%%%%%%%%%%%%%%%%%%%%%%%%%%%%%%%%
% ---  ---  ---  ---  ---  ---  ---  ---  --- Intro ---  ---  ---  ---  ---  ---  ---  ---  ---
%%%%%%%%%%%%%%%%%%%%%%%%%%%%%%%%%%%%%%%%%%%%%%%%%%%%%%%%%%%%%%%%%%%%%%%%%%%%%%%%%%%%%%%%%%%%%%%
%%%%%%%%%%%%%%%%%%%%%%%%%%%%%%%%%%%%%%%%%%%%%%%%%%%%%%%%%%%%%%%%%%%%%%%%%%%%%%%%%%%%%%%%%%%%%%%
\section{Introduction}
\label{sec:intro}

% Introduce the task - keypoints annotations much much cheaper, pose estimators are advanced and could provide good starting point for segmentation, BMP loop
% SAM - SAM revolutionized the field with foundational model and training on huge dataset with diverse images and videos, but SAM has no notion of semantics
% We are leveraging power of SAM and its generalization and add semantic to it to create SAM-pose2seg
% The semantic is added in two ways. (1) The decoder is fine-tuned on human-only examples and as such learns to predict human segmentation masks in a robust way. (2) The point encoder is trained on human body keypoints instead of randomly sampled points. That makes the encoder more aligned with the target task.
% Freezing backbone give our model generalization abilities while training on generated keypoints gives it robustness to localization noise.
% The main contribution of this paper is SAM-pose2seg: a fine-tuned SAM 2.1 model with training adapted for pose-guided human instance segmentation task. It achieves SOTA on predicted-pose-guided human instance segmentation as well as gt-pose-guided human instance segmentation.

A standard approach to human instance segmentation is to use a detector that predicts instance masks directly.
However, in crowded scenes with heavy occlusion, these detectors often fail to separate overlapping instances.
In contrast, human pose estimators are more robust in these conditions and produce structured keypoints that are easier to annotate and more stable under clutter.

Keypoint annotations are also significantly cheaper than per-pixel segmentations, making pose a practical intermediate representation for instance segmentation.
In particular, human keypoints can serve as effective prompts for segmentation models.
The task of \textit{pose-guided human instance segmentation} takes either ground-truth or detected keypoints as input and outputs a segmentation mask for each instance.

This task was introduced with the OCHuman dataset \cite{pose2seg} and the pose2seg model and was explored in many models \cite{pose2seg, pose2instance, PosePlusSeg, PoSeg, MultiPoseSeg, PoseSegTTA} since then.
More recently, it has been used in the self-improving BMP loop (BBoxMaskPose, \cite{BMPv1}), where pose-guided segmentation plays a key role in resolving multi-body ambiguity in heavily occluded scenes.

The Segment Anything Model (SAM, \cite{SAM1}) introduced a general segmentation framework trained on a large and diverse dataset of masks and images.
SAM 2 \cite{SAM2} extended this with large-scale video training.
While SAM shows strong generalization and has revolutionized prompted segmentation, it lacks semantic understanding and is not specialized for any class, such as humans.

We build on SAM’s generalization and introduce a method to adapt it for pose-guided human instance segmentation.
Our model, SAM-pose2seg, incorporates semantic information into SAM through two main modifications.
First, we fine-tune the decoder on human-only segmentation masks to specialize the predictions.
Second, we replace SAM’s random point prompts with body keypoints during training.
This aligns the prompt encoder and mask decoder more closely with the target task.

% We keep the SAM backbone frozen to retain its generalization, while training on detected keypoints improves robustness to localization noise.

SAM-pose2seg is a pose-guided variant of SAM 2.1, fine-tuned for human instance segmentation.
It introduces semantic specialization through decoder fine-tuning and aligns the prompt encoder with the task by training on human keypoints.
This design enables robust segmentation from both the predicted pose and the ground-truth.
In the remainder of the paper, we show that SAM-pose2seg achieves state-of-the-art performance on pose-guided human instance segmentation benchmarks and analyze design choices through ablation studies.

% SAM is an effective tool for generating segmentation masks based on pose keypoints. In this work, we aim to find the best way to use SAM 2.1 for human segmentation mainly in cases where parts of the body are occluded. To do this, we explore improvements in prompting methods and fine-tune the model to better match our task. SAM-pose2seg will be used to improve the iterative pose refinement in the BboxMaskPose~\cite{bboxmaskpose} loop, though it also achieves leading positions in pose-to-segmentation task.

%%%%%%%%%%%%%%%%%%%%%%%%%%%%%%%%%%%%%%%%%%%%%%%%%%%%%%%%%%%%%%%%%%%%%%%%%%%%%%%%%%%%%%%%%%%%%%%
%%%%%%%%%%%%%%%%%%%%%%%%%%%%%%%%%%%%%%%%%%%%%%%%%%%%%%%%%%%%%%%%%%%%%%%%%%%%%%%%%%%%%%%%%%%%%%%
% ---  ---  ---  ---  ---  ---  ---  ---  --- Intro ---  ---  ---  ---  ---  ---  ---  ---  ---
%%%%%%%%%%%%%%%%%%%%%%%%%%%%%%%%%%%%%%%%%%%%%%%%%%%%%%%%%%%%%%%%%%%%%%%%%%%%%%%%%%%%%%%%%%%%%%%
%%%%%%%%%%%%%%%%%%%%%%%%%%%%%%%%%%%%%%%%%%%%%%%%%%%%%%%%%%%%%%%%%%%%%%%%%%%%%%%%%%%%%%%%%%%%%%%
\section{Related Work}
\label{sec:related}

\noindent
\textbf{Detection-Based Segmentation:}
The most direct approach to human instance segmentation uses a detector \cite{coDETR, YOLOX, YOLOv3, RTMDet, ViTDet, HRNet, ConvNeXt, HRNet}.
A detector takes an image as input and outputs instance masks and class labels.
Detectors work well in scenes similar to the training data, but struggle when instances overlap heavily.
Under severe occlusion, they often merge multiple people into one mask, or assign different body parts to different instances.
Tuning hyperparameters (e.g., non-maximum suppression) can reduce this, but increases false positives.

\noindent
\textbf{General Prompted Segmentation}
requires an image and a human or automatically generated prompt.
A prominent example is the SAM family \cite{SAM1, SAM2}, trained on large image and video datasets with human-in-the-loop supervision.
These models generalize well, but lack semantic understanding, making the task inherently ambiguous.
As a result, they often segment skin, face, hair, clothing, or body parts instead of the full human instance.

\noindent
\textbf{Semantics-Aware Prompted Segmentation:}
Several works \cite{GroundedSAM, SAMWise, SA-SAM, OpenWorldSAM} inject semantics into SAM.
They take an image and a text prompt as input and detect and segment instances.
This setting is often referred to as \textit{open-vocabulary} or \textit{zero-shot} segmentation.
While these models capture semantics and segment entire instances, they do not provide localized cues and offer no control over which instance is chosen.
Moreover, they are unsuitable for iterative loops such as BMP \cite{BMPv1}.

% \cite{pose2seg, pose2instance, PosePlusSeg, PoSeg, MultiPoseSeg, PoseSegTTA, CrowdSAM, ParsingRCNN, ttattg}
\noindent
\textbf{Pose-Aware Instance Segmentation}
is the most specific form of prompted segmentation.
The model takes an image and a detected or ground-truth human pose and segments the corresponding person.
These methods specialize in humans, sacrificing the generality of open-vocabulary or generic segmenters for robustness and precise, localized control.
% They can be divided into several groups.
%
Test-time optimization methods \cite{ttattg, PoseSegTTA} refine predictions at inference without training.
While training-free, they are computationally expensive and slow at test time.
Standard pose-guided human segmentation methods \cite{pose2body, pose2instance, pose2seg, PosePlusSeg, PoSeg, MultiPoseSeg, ParsingRCNN} rely on small training datasets and thus generalize poorly, especially in crowded scenes.
The closest related work is CrowdSAM \cite{CrowdSAM}, which builds a framework around SAM to automatically annotate bounding boxes in crowds.
CrowdSAM uses SAM and DINO \cite{DINOv2} as external tools and optimizes prompts, similar to \cite{BMPv1}.
In contrast, our SAM-pose2seg is end-to-end and, when used in an iterative loop, outperforms CrowdSAM, as shown in \cite{BMPv1}.

\noindent
\textbf{Datasets:}
There are not many datasets with annotated human instance segmentation masks and human poses.
One of the first to offer this kind of data was COCO \cite{COCO} and it became a standard for pose-guided segmentation training.
COCO has two problems.
First, it does not focus on overlapping instances, which is now the most challenging scenario (see \cite{BMPv1}).
Second, the human pose is annotated only for ``large`` instances, so the models are not trained or evaluated on small background persons.
Overlapping instances were addressed in OCHuman dataset \cite{pose2seg} but it is too small for training and contains only validation and testing sets.
There are datasets such as \cite{MPII, AIC, CrowdHuman, CrowdPose, CIHP} focusing on analysis of human body in crowded scenes, but none of these offer both segmentation and pose annotations.
In this work, we worked with COCO \cite{COCO} and CIHP \cite{CIHP} for training and OCHuman \cite{pose2seg} for evaluation. For details about used datasets, see \cref{sec:experiments}.

\noindent
\textbf{Iterative Methods:}
Pose-guided instance segmentation has been used in iterative pipelines \cite{IterDet, BMPv1}.
Our work builds on BMP \cite{BMPv1}, where instance segmentation is prompted by detected human keypoints.
They use unmodified SAM 2 with a complex prompt selection strategy.
We also build on SAM 2, but adapt it to this task.
Our SAM-pose2seg is fine-tuned for human instances, removing the need for complex prompt selection.
Compared to SAM 2 in BMP, SAM-pose2seg achieves better performance, with higher robustness and lower complexity.

%%%%%%%%%%%%%%%%%%%%%%%%%%%%%%%%%%%%%%%%%%%%%%%%%%%%%%%%%%%%%%%%%%%%%%%%%%%%%%%%%%%%%%%%%%%%%%%
%%%%%%%%%%%%%%%%%%%%%%%%%%%%%%%%%%%%%%%%%%%%%%%%%%%%%%%%%%%%%%%%%%%%%%%%%%%%%%%%%%%%%%%%%%%%%%%
% ---  ---  ---  ---  ---  ---  ---  ---  --- Intro ---  ---  ---  ---  ---  ---  ---  ---  ---
%%%%%%%%%%%%%%%%%%%%%%%%%%%%%%%%%%%%%%%%%%%%%%%%%%%%%%%%%%%%%%%%%%%%%%%%%%%%%%%%%%%%%%%%%%%%%%%
%%%%%%%%%%%%%%%%%%%%%%%%%%%%%%%%%%%%%%%%%%%%%%%%%%%%%%%%%%%%%%%%%%%%%%%%%%%%%%%%%%%%%%%%%%%%%%%

\section{Method}
\label{sec:method}
%%%% Some ideas to rename methods MaxVis, MaxSpread, MaskRefine and MaskRefine-Pose %%%%
% Idea: MaxVis is selecting top-n visible and is maximizing visible keypoints
% --> TopVis-n, TopV-n, MaxVis-n, VMax, max(V)

% Idea: MaxSpread is selecting the most spread keypoints and is maximizing keypoints spread
% --> MostSpread-n, TopS-n, MaxSpread-n, SMax, max(S)

% Idea: MaskRefine is selecting points from gt/dt error and is minimizing prediction error
% --> MaskRefine-n, ErrSample-n, MaskIter-n, MinErr-n, EMin, min(E)

% Idea: MaskRefine-Pose is good in a sense that it (1) builds on MaskRefine and (2) bring semantics
% --> PoseMaskRefine-n, PoseErrSample-n, PoseMaskIter-n, PoseMinErr-n, Pose-EMin, min(EP), min(P)

%%%%%%%%%%%%%%%%%%%%%%%%%%
%%%%  Section outline %%%%

\subsection{SAM}
Segment Anything (SAM) is a powerful, robust tool with great generalization capacity. Its robustness to noise and point prompting capability offers a highly optimal foundation for our task. However, standard prompting techniques often struggle in occluded scenarios, where detected keypoints lack unequivocal visibility information, making SAM unreliable.

Our goal is to adapt SAM (specifically, \textit{SAM 2.1}) for pose-guided segmentation while retaining its ability to generalize. We introduce minimal modifications to the encoder structure to tailor the model specifically for this task.

\subsection{Fine-tuning}
We fine-tuned the \textit{SAM~2.1 Hiera Base Plus} model using the official Meta training script, focusing specifically on the point selection mechanism. To preserve the generalization power of the backbone—trained on massive, diverse dataset—and to accelerate training, we froze the image encoder. Only the prompt encoder and mask decoder were optimized. No architectural changes were made to the mask decoder; however, it was also a part of the training process, as the prompt encoder's only learned parameters are the weights for positive, negative, and bounding box embeddings. We adopted the sampling strategy \textit{PoseMaskRefine}.

\paragraph{Method MaskRefine (Default Training Strategy).}
Meta's default training procedure serves as a strong baseline, which only builds on the ground-truth (GT) masks. It does not rely on pre-defined semantic or pose-based points, instead, they are sampled dynamically: the first is drawn uniformly from the ground-truth mask, while the remaining seven are sampled from the error region between the GT mask and the previous prediction (with a small probability, sampling is done from the GT mask instead). This totals eight points per iteration. By conditioning the model on both the previous logits and the newly sampled point, this method enables iterative refinement without dependence on explicit semantic cues.

\paragraph{Method PoseMaskRefine (Pose-Guided Refinement).}
Building on the strong performance of MaskRefine, we introduce PoseMaskRefine to meet our specific task requirements. Here, the initial point is not sampled uniformly but is selected as an available pose keypoint with the highest visibility, unless no keypoint is available. In subsequent iterations, the remaining seven points are preferentially sampled from pose keypoints located within the current error region; with a small probability, sampling follows the original MaskRefine strategy and selects points uniformly from the ground-truth mask instead. If no such keypoints exist, sampling reverts to uniform selection from the error region. By incorporating pose cues while preserving the iterative, error-driven nature of MaskRefine, this strategy consistently outperformed the default approach.

The MaskRefine strategy effectively acts as hard negative mining; because samples are selected based on local error, boundary and occluded regions receive strong supervision. We attempted to simplify this by eliminating the iterative correction process in favor of a pure keypoint-based prediction to reflect the inference prompting (see \cref{sec:experiments-ablation}), but this yielded no significant improvement.

\begin{figure}[tb]
    \centering

    % --- COLUMN 1: Left Third (Input) ---
    % Trim: Removes the right 2/3 (approx 0.666)
    \begin{subfigure}{0.32\linewidth}
        \centering
        \adjincludegraphics[width=\linewidth, trim={0 0 {0.666\width} 0}, clip]{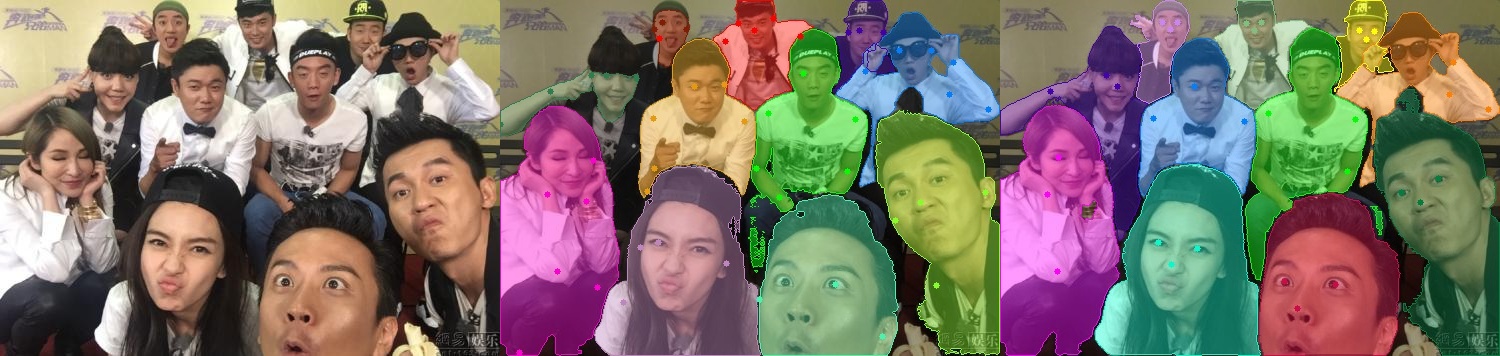}
        \caption*{Input}
    \end{subfigure}
    \hfill
    % --- COLUMN 2: Center Third (Base SAM 2.1) ---
    % Trim: Removes left 1/3 (0.333) and right 1/3 (0.333)
    \begin{subfigure}{0.32\linewidth}
        \centering
        \adjincludegraphics[width=\linewidth, trim={{0.333\width} 0 {0.333\width} 0}, clip]{imgs/SAM-pose2seg/combined_0037472_new.jpg}
        \caption*{SAM 2.1}
    \end{subfigure}
    \hfill
    % --- COLUMN 3: Right Third (SAM-pose2seg) ---
    % Trim: Removes left 2/3 (approx 0.666)
    \begin{subfigure}{0.32\linewidth}
        \centering
        \adjincludegraphics[width=\linewidth, trim={{0.666\width} 0 0 0}, clip]{imgs/SAM-pose2seg/combined_0037472_new.jpg}
        \caption*{SAM-pose2seg}
    \end{subfigure}
    \vspace{-0.5em}
    \caption{
    \textbf{Improved handling of \textcolor{blue}{incomplete segmentation}}, with SAM-pose2seg correctly segmenting full human instances. More examples in Supplementary.
    \vspace{-1em}
    }
    \label{fig:SAM-pose2seg-probl-incompsegm}
\end{figure}

Fine-tuning with PoseMaskRefine makes the model significantly less sensitive to common errors, such as segmenting clothing only or omitting body parts (see \cref{fig:SAM-pose2seg-probl-incompsegm}). 

\subsection{Prompting}
Effective prompting is crucial for interacting with SAM, yet pose keypoints do not perfectly align with the correction-based prompting concept inherent to SAM's training. Therefore, our goal was to find the most effective way to pose prompting while sticking to the SAM's structure. 

Firstly, we incorporate the ProbPose \cite{ProbPose} \emph{visibility} metric as a reliable criterion for keypoint selection. This metric serves as a confidence measure for each keypoint, indicating whether the respective body part is observable in the picture and enabling the exclusion of keypoints that are likely occluded or noisy.

\begin{figure}[tb]
\centering

% In the preamble (or right before the figure)
\definecolor{coco}{RGB}{52, 101, 164}      % blue
\definecolor{ochuman}{RGB}{204, 0, 0}      % red
\definecolor{cihp}{RGB}{78, 154, 6}        % green

\begin{tikzpicture}
% ---------- LEFT AXIS: COCO ----------
\begin{axis}[
    width=0.88\linewidth,
    height=0.6\linewidth,
    xlabel={\# keypoints},
    ylabel={COCO AP},
    ylabel style={color=coco},
    yticklabel style={color=coco},
    xmin=2.8, xmax=17.5,
    ymin=20, ymax=46,
    xtick={3,6,8,12,17},
    ytick={0,5,...,50},
    grid=both,
    major grid style={line width=0.3pt, draw=gray!35},
    minor grid style={line width=0.2pt, draw=gray!20},
    minor tick num=1,
    tick align=outside,
    axis y line*=left,
    axis x line*=bottom,
]

% % --- COCO (color 1) ---
\addplot+[very thick, mark=none, color=coco, mark options={draw=coco, fill=coco}, mark size=1.6pt] coordinates {
    (3,24.5) (6,37.7) (8,38.2) (12,31.6) (17,13.2)
};
\addplot+[very thick, dashed, color=coco, mark color=coco, mark=none, mark options={draw=coco, fill=coco}, mark size=1.6pt] coordinates {
    (3,36.2) (6,41.2) (8,41.1) (12,38.5) (17,35.7)
};

\end{axis}

% --- CIHP (green) ---
% ---------- RIGHT AXIS: CIHP ----------
\begin{axis}[
    width=0.88\linewidth,
    height=0.6\linewidth,
    xmin=2.8, xmax=17.5,
    ymin=30, ymax=75,
    ytick={10,20,...,80},
    ylabel={CIHP AP},
    ylabel style={color=cihp},
    yticklabel style={color=cihp},
    axis y line*=right,
    axis x line=none,
    tick align=outside,
]

% --- OCHuman (color 2) ---
% \addplot+[very thick, mark=square*, color=ochuman, mark size=1.6pt] coordinates {
%     (3,21.5) (6,29.4) (8,30.7) (12,29.7) (17,10.9)
% };
% \addplot+[very thick, dashed, mark=square*, color=ochuman, mark options={draw=ochuman, fill=ochuman}, mark size=1.6pt] coordinates {
%     (3,25.5) (6,29.5) (8,29.9) (12,28.3) (17,27.4)
% };

% --- CIHP (color 3) ---
\addplot+[very thick, mark=none, color=cihp, mark options={draw=cihp, fill=cihp}, mark size=1.6pt] coordinates {
    (3,60.1) (6,66.4) (8,65.8) (12,50.3) (17,11.0)
};
\addplot+[very thick, dashed, mark=none, color=cihp, mark options={draw=cihp, fill=cihp}, mark size=1.6pt] coordinates {
    (3,68.3) (6,71.6) (8,71.1) (12,69.1) (17,67.9)
};

\node[
    anchor=south west,
    font=\scriptsize,
    fill=white,
    text=DarkerGray,
    fill opacity=0.7,
    text opacity=1,
    inner sep=4pt
] at (axis description cs:0.1,0.02)
{
\begin{tabular}{@{}l@{}}
\tikz{\draw[very thick] (0,0.5)--(0.8,0.5);} SAM 2.1 + MaxVis\\
\tikz{\draw[very thick, dashed] (0,0.5)--(0.8,0.5);} SAM 2.1 + MaxSpread\\
\end{tabular}
};

\end{axis}

\end{tikzpicture}

\caption{
\textbf{Keypoint selection methods on SAM 2.1}.
Prompting methods MaxVis (full) and MaxSpread (dashed) on \textcolor{coco}{COCO} and \textcolor{cihp}{CIHP} datasets.
6 keypoints is the best for both methods.
MaxSpread outperforms MaxVis as shown in \cite{BMPv1}.
% AP vs. number of keypoints for two keypoint selection strategies. Colors denote datasets; solid lines are MaxVis (final choice), dashed lines are MaxSpread.
}
\label{fig:selection-method-plot-SAM}
\end{figure}
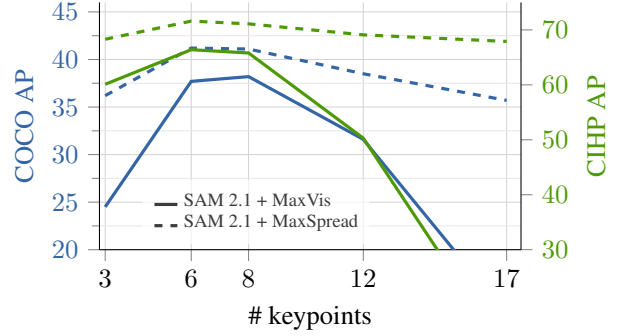

For the base model, selecting six keypoints based on both visibility and distance proved most stable (see \cref{fig:selection-method-plot-SAM}), as this provided sufficient variability to recognize the whole body without the detriment caused by exceeding eight points.

\begin{figure}[tb]
    \centering

    % --- COLUMN 1: Left Third (Input) ---
    % Trim: Removes the right 2/3 (approx 0.666)
    \begin{subfigure}{0.32\linewidth}
        \centering
        \adjincludegraphics[width=\linewidth, trim={0 0 {0.666\width} 0}, clip]{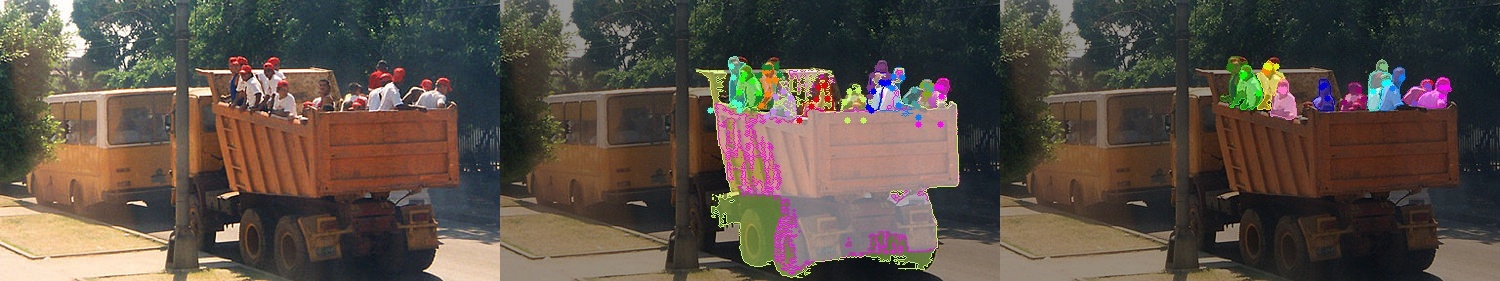}
        \caption*{Input}
    \end{subfigure}
    \hfill
    \begin{subfigure}{0.32\linewidth}
        \centering
        \adjincludegraphics[width=\linewidth, trim={{0.333\width} 0 {0.333\width} 0}, clip]{imgs/SAM-pose2seg/combined_000000259571_new.jpg}
        \caption*{Base SAM 2.1}
    \end{subfigure}
    \hfill
    \begin{subfigure}{0.32\linewidth}
        \centering
        \adjincludegraphics[width=\linewidth, trim={{0.666\width} 0 0 0}, clip]{imgs/SAM-pose2seg/combined_000000259571_new.jpg}
        \caption*{SAM-pose2seg}
    \end{subfigure}
    \vspace{-0.5em}
    \caption{
    \textbf{Improved handling of \textcolor{cyan}{partially occluded poses}.} More examples in Supplementary.
    }
    \vspace{-1em}
    \label{fig:SAM-pose2seg-probl-partoccl}
\end{figure}

However, across all fine-tuning experiments, PoseMaskRefine consistently yielded the best generalization and enabled a significant simplification of our inference strategy. Its flexibility allows us to replace complex heuristics with a simpler approach, reducing the risk of selecting occluded points (see \cref{fig:SAM-pose2seg-probl-partoccl}). We now select only the \textbf{3 keypoints with the highest visibility scores}. This simplified strategy improved accuracy across all observed datasets using detected keypoints. 

Note: Consequently, we employ distinct selection methods for the base SAM 2.1 and SAM-pose2seg in figures unless stated otherwise to compare the optimal performance of each version.

\section{Experiments}
\label{sec:experiments}

%%%%%%%%%%%%%%%%%%%%%%%%%%%%%%%%%%%%%%%%%%%%%%%%%%%%%%%%%%%%%%%%%%%%%%%%%%%%%%%%%%%%%%%%%%%%%%%
% ---  ---  ---  ---  ---  ---  ---  ---  ---  ---  ---  ---  ---  ---  ---  ---  ---  ---  ---
\subsection{Implementation Details}
\label{sec:experiments-details}

%%%% Subchapter Outline %%%% --> I would have a small titles just as bolt text at the beginning of each paragraph, similarly what I do in Related work (eg in BMPv1 -- https://mirapurkrabek.github.io/BBox-Mask-Pose/static/pdfs/BBoxMaskPose.pdf)
% **Data** Which data we used - combined training on COCO and CIHP. For both, we generated detected pose keypoints for ALL instances (including small ones that are usually ignored for pose estimation) so that our model is robust to localization noise
% **Architecture** -- mostly intact, but says which model we used etc
% **Training details** -- freezing backbone, LR, how many epochs...

\paragraph{Data.} For training, we used the COCO and CIHP datasets together with ProbPose \cite{ProbPose} keypoints to ensure a wide range of human poses and occlusion conditions, mirroring their main use at inference time
Training on COCO alone is not sufficient due to the pronounced domain shift relative to CIHP and OCHuman, both of which emphasize multi-person and heavily occluded scenarios.
We generated detected pose keypoints for all instances (including small ones that are usually ignored for pose estimation) so that our model is robust to localization noise.

\paragraph{Architecture and Training Details.}
We fine-tuned the \textit{SAM~2.1 Hiera Base Plus} checkpoint. We trained only the prompt and mask decoders, while the image encoder (backbone) remained frozen. The configuration file, \texttt{SAM 2.1\_hiera\_b+\_MOSE\_finetune.yaml}, was taken from the SAM~2 repository.\footnote{Available at \url{https://github.com/facebookresearch/sam2}} All hyperparameters not mentioned here were left unchanged. The number of frames was set to 1. We trained for 15 epochs, though the model converged earlier; performance changes were marginal when varying epochs between 10 and 30. The probability of bounding box usage (\texttt{prob\_to\_use\_box\_input\_for\_train}) was set to 0.0, while the probability of point usage (\texttt{prob\_to\_use\_pt\_input\_for\_train}) was set to 1.0.
% \begin{table}[tb]
% \caption{
%     \textbf{Evaluation of the various fine-tuning methods.} Each fine-tuning method differs in dataset used keypoint selection method during the training and inference. "Keypoint selection" row is in format \textit{selection during training} or \textit{selection during training/selection during inference} if inference method differs.
%     }
% \label{tab:training-method-evals}
% \centering
% \renewcommand{\arraystretch}{1.6}
% \small
% \begin{tabular}{lcccc}
% \toprule
% Dataset & \makecell{Keypoint\\selection} & 
% \makecell{COCO\\val AP} &
% \makecell{OCHuman\\test AP} &
% \makecell{CIHP\\val AP} \\
% \midrule
% \rowcolor{gray!15}
% base & MaxVis$_6$ & 37.7 & 29.4 & 66.4 \\
% COCO & MaxVis$_6$ & 41.2 & 27.0 & 62.2 \\
% \rowcolor{gray!15}
% base & MaxSpread$_6$ & 41.2 & 29.5 & 71.6 \\
% COCO & MaxSpread$_6$ & 42.6 & 26.1 & 64.3 \\
% \makecell{COCO+\\CIHP} & MaxSpread$_6$ & 43.3 & 29.3 & 67.7 \\
% COCO & \makecell{MaskRefine/\\MaxVis$_3$} & 43.7 & 29.8 & 68.9 \\
% \makecell{COCO+\\CIHP} & \makecell{MaskRefine/\\MaxVis$_3$} & 43.7 & 34.1 & 71.7 \\
% \makecell{COCO+\\CIHP} & \makecell{PoseMaskRefine$_1$/\\MaxVis$_3$} & 44.5 & 34.5 & 72.7 \\
% \makecell{COCO+\\CIHP} & \makecell{PoseMaskRefine/\\MaxVis$_3$} & \textbf{44.6} & \textbf{34.7} & \textbf{72.7} \\
% \bottomrule
% \end{tabular}
% \end{table}

\begin{table}[tb]
\centering
\small
\begin{tabular}{ccccc | ccc}
\toprule
\rotatebox[origin=c]{90}{COCO} & \rotatebox[origin=c]{90}{CIHP} & \rotatebox[origin=c]{90}{\makecell{train\\selection}} & \rotatebox[origin=c]{90}{\makecell{test\\selection}} & \rotatebox[origin=c]{90}{\# points} &
\makecell{COCO\\AP} &
\makecell{OCH\\AP} &
\makecell{CIHP\\AP} \\
\midrule

% \multicolumn{2}{c}{baseline}  & MaxVis & 6 & 37.7 & 29.4 & 66.4 \\
\xmark & \xmark            & mV & mV & 6 & 37.7 & 29.4 & 66.4 \\
\xmark & \xmark            & mS & mS & 6 & 41.2 & 29.5 & 71.6 \\
\midrule
\cmark & \xmark            & mV & mV & 6 & 41.2 & 27.0 & 62.2 \\
\cmark & \xmark            & mS & mS & 6 & 42.6 & 26.1 & 64.3 \\
\cmark & \cmark            & mS & mS & 6 & 43.3 & 29.3 & 67.7 \\
\midrule
\cmark & \xmark           & mR & mV & 3 & 43.7 & 29.8 & 68.9 \\
\cmark & \cmark           & mR & mV & 3 & 43.7 & 34.1 & 71.7 \\
\cmark & \cmark          & P1mR & mV & 3 & 44.5 & 34.5 & 72.7 \\
\cmark & \cmark          & PmR & mV & 3 & \textbf{44.6} & \textbf{34.7} & \textbf{72.7} \\
\bottomrule
\end{tabular}
\caption{
    % \textbf{Evaluation of the various fine-tuning methods.} Each fine-tuning method differs in dataset used keypoint selection method during the training and inference. "Keypoint selection" row is in format \textit{selection during training} or \textit{selection during training/selection during inference} if inference method differs.
    \textbf{Ablation study: fine-tuning recipes.}
    Keypoint selection methods are MaxVis~(mV), MaxSpread~(mS), MaskRefine~(mR), PoseMaskRefine~(PmR) and Pose1MaskRefine~(P1mR).
    First two rows are baselines SAM 2.1 without any fine-tuning.
    For mV and mS methods, the optimal \# points is 6, for mR-based methods it is 3.
    The best point method is PmR with 3 points.
    Training on both COCO and CIHP is crucial for generalization to unseen OCHuman.
    }
\label{tab:training-method-evals}
\end{table}

%%%%%%%%%%%%%%%%%%%%%%%%%%%%%%%%%%%%%%%%%%%%%%%%%%%%%%%%%%%%%%%%%%%%%%%%%%%%%%%%%%%%%%%%%%%%%%%
% ---  ---  ---  ---  ---  ---  ---  ---  ---  ---  ---  ---  ---  ---  ---  ---  ---  ---  ---
\subsection{SOTA Comparison}
\label{sec:experiments-SOTA}

The proposed \textit{SAM-pose2seg} model represents a strong all-round solution for the pose-to-segmentation task. Across a wide range of datasets and evaluation settings, it consistently achieves competitive or superior performance. In particular, SAM-pose2seg demonstrates robust behavior under both detected and ground-truth pose inputs, making it well suited for practical deployment scenarios.

Using detected ProbPose keypoints, we achieve \textbf{44.6 AP} on COCO val2017, \textbf{60.3 AP} on COCOPersons (COCO val2017 excluding small instances), \textbf{34.7 AP} on OCHuman test, and \textbf{72.7 AP} on CIHP val. These results demonstrate strong generalization across datasets with varying levels of occlusion and pose complexity, with particularly strong performance on the challenging OCHuman benchmark.

When evaluated with ground-truth poses, SAM-pose2seg further improves performance, reaching \textbf{61.6 AP} on COCOPersons, \textbf{70.0 AP} on OCHuman test, and \textbf{69.5 AP} on OCHuman val. For ground-truth keypoints, we apply spread-based keypoint selection, as their binary visibility annotations do not allow for visibility-based ranking.

\begin{table}[tb]
\centering
\setlength{\tabcolsep}{2pt} % reduce column spacing
\scriptsize
\small
\begin{tabular}{@{}l|ccccc@{}}
\toprule
Model (Prompting) & \makecell{COCO\\val AP} & \makecell{COCO*\\val AP} & \makecell{OCHuman\\test AP} \\
\midrule
HQNet R-50 & - & - & 31.1 \\
Pose2Seg & - & 55.5 & 23.8 \\
ExPoSeg$^+$ & - & 61.9 & 26.8 \\
Occlusion C\&P$^+$ & - & - & 28.3 \\
Crowd-SAM$^+$ & 22.0 & - & 31.4 \\
MultiPoseSeg & - & 56.3 & - \\
SAM 2.1 & 41.2 & 56.0 & 29.5 \\
\rowcolor{green!15}
SAM-pose2seg & 44.6 & 60.3 & 34.7 \\
\bottomrule
\end{tabular}
\caption{\textbf{Evaluation of the base SAM model and SAM-pose2seg} on \textbf{detected ProbPose keypoints}.
The improvement is most visible on the OCHuman test set. Prompting methods are explained in \cref{sec:keypoint-selection}. COCO* val AP is evaluated without Small category persons (no ground-truth poses are available there) for comparison with Pose2Seg. Models labeled with $^+$ estimate either masks or report detection AP. Every result except for SAM 2.1 was achieved on a different set of detected keypoints.}
\label{tab:detect-SAM-pose2seg-res}
\end{table}

\begin{table}[tb]
\centering
\setlength{\tabcolsep}{2pt} % reduce column spacing
\scriptsize
\small
\begin{tabular}{@{}l|cccc@{}}
\toprule
Model (Prompting) & \makecell{COCO*\\val AP} & \makecell{OCHuman\\test AP} & \makecell{OCHuman\\val AP} \\
\midrule

Pose2Seg & 58.2 & 55.2 & 54.4 \\ 
base SAM 2.1 & 57.1 & 70.1 & 70.2 \\
\rowcolor{green!15}
SAM-pose2seg  & 61.6 & 70.0 & 69.5 \\

\bottomrule
\end{tabular}
\caption{\textbf{Evaluation of the base SAM model and SAM-pose2seg} AP on the \textbf{ground-truth pose keypoints}. MaxVis strategy does not make sense as the visibility classification is binary here, we therefore use selection by distance to optimize, while sorting out all keypoints with visibilty set to zero. *COCO val AP is only evaluated without Small category persons, as human pose annotations are missing there. MaxSpread$_3$ keypoint selection is used for SAM-pose2seg.
}
\label{tab:gt-SAM-pose2seg-res}
\end{table}

%%%%%%%%%%%%%%%%%%%%%%%%%%%%%%%%%%%%%%%%%%%%%%%%%%%%%%%%%%%%%%%%%%%%%%%%%%%%%%%%%%%%%%%%%%%%%%%
% ---  ---  ---  ---  ---  ---  ---  ---  ---  ---  ---  ---  ---  ---  ---  ---  ---  ---  ---
\subsection{Ablation Study}
\label{sec:experiments-ablation}
\subsubsection{Correction Points In Training}

We further review the correction process to prove why our model converges to the optimal usage of three keypoints during inference. Crucially, as the model adapts to human segmentation, it rarely requires all seven correction points. In practice, the fine-tuned \textit{SAM-pose2seg} model often predicts a correct mask after the first sampled keypoint. When comparing ground-truth masks to masks in each of the correction iteration, it was discovered that the differences changes between the mask in each iteration and their IoU with the ground-truth mask are small -- mean IoU on a small subset during before any correction iterations is already 81.0 \% and only goes up by 6.4 percentage points after the last iteration.
 This is corroborated by our \textit{Pose1MaskRefine} experiment, where only the first point is pose-derived, and subsequent correction points are sampled uniformly as in the default method. The marginal difference in AP between PoseMaskRefine$_1$ and the full PoseMaskRefine confirms that the first keypoint is the primary driver of performance (see \cref{tab:training-method-evals}).
 
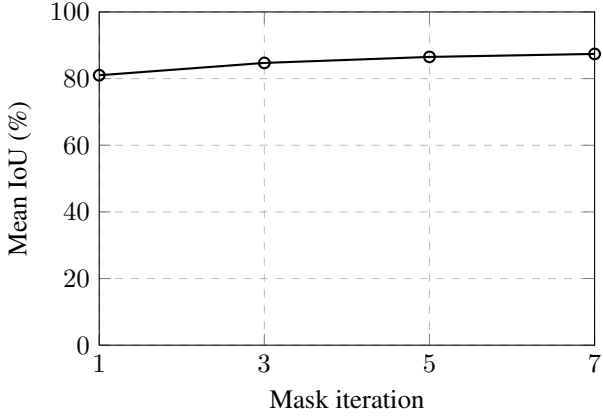
\begin{figure}[tb]
    \centering
    \begin{tikzpicture}
        \begin{axis}[
            width=\linewidth,
            height=6cm,
            xlabel={Mask iteration},
            ylabel={Mean IoU (\%)},
            xmin=1, xmax=7,
            ymin=0, ymax=100,
            xtick={1,3,5,7},
            ytick={0,20,40,60,80,100},
            grid=both,
            major grid style={dashed},
        ]
            \addplot[
                mark=o,
                thick
            ] coordinates {
                (1,81.0)
                (3,84.7)
                (5,86.5)
                (7,87.4)
            };
        \end{axis}
    \end{tikzpicture}
    \caption{
    \textbf{Importance of correction points in the default SAM training method MaskRefine.} Comparison on a small COCO + CIHP training subset of mean IoU of the ground-truth mask. The first iteration contains only one point, any other \emph{i}-th iteration is prompted by \emph{i} keypoints and a correction mask. It seems that the refinement mostly causes minor changes in the overall mask shape.
    }
    \label{fig:mask_correction_iteration_iou}
\end{figure}

\subsubsection{Prompting -- Keypoint Selection}
\label{sec:keypoint-selection}
We evaluate two prompting strategies, MaxVis and MaxSpread, which differ in keypoint selection and interaction with ground-truth (GT) masks during inference and training (yet we stuck with PoseMaskRefine in the case of training).

\paragraph{Method MaxVis$_n$ (Visibility-Based Keypoint Selection).}
This method selects the top $n$ keypoints solely based on \textit{visibility} scores, imposing no additional spatial, semantic, or structural constraints.

\paragraph{Method MaxSpread$_n$ (Distance- and Visibility-Based Keypoint Selection).}
Originally introduced in Bbox-Mask-Pose~\cite{BMPv1} (specifically MaxSpread$_6$), this method extends visibility-based selection with a spatial diversification heuristic inspired by \emph{k-means++} \cite{kmeans++}. The first keypoint is chosen for maximum visibility; subsequent points are selected to maximize distance from previous ones. To reduce redundancy, number of used eye and nose keypoints is limited to one, and low-visibility points are excluded. This strategy generally outperforms MaxVis on the default model and is preferred for binary visibility classification.

MaxSpread$_6$ proved most effective for pose-to-segmentation, remaining consistent after incorporating ProbPose visibility. For the base model, spatially distributed keypoints provide richer cues than simple visibility selection, which often overrepresents facial points and causes segmentation ambiguity (e.g., face vs. full body). However, after fine-tuning with MaskRefine and PoseMaskRefine, the performance gap between MaxSpread$_n$ and MaxVis$_n$ becomes negligible.

% \begin{table}[tb]
% \centering
% \begin{tabular}{l | ccc}
% \toprule
% \# keypoints & \makecell{COCO\\val AP} & \makecell{OCHuman\\test AP} & \makecell{CIHP\\val AP} \\
% \midrule
% \multicolumn{4}{l}{MaxVis} \\
% 3 & 24.5 & 21.5 & 60.1 \\
% 6 & 37.7 & 29.4 & \textbf{66.4} \\
% 8 & \textbf{38.2} & \textbf{30.7} & 65.8 \\
% 12 & 31.6 & 29.7 & 50.3 \\
% 17 & 13.2 & 10.9 & 11.0 \\

% \multicolumn{4}{l}{MaxSpread} \\
% 3 & 36.2 & 25.5 & 68.3 \\
% 6 & \textbf{41.2} & \textbf{29.5} & \textbf{71.6} \\
% 8 & \textbf{41.1} & \textbf{29.9} & 71.1 \\
% 12 & 38.5 & 28.3 & 69.1 \\
% 17 & 35.7 & 27.4 & 67.9 \\
% \bottomrule
% \end{tabular}
% \caption{
%     \textbf{Comparison of keypoint selection methods on base model.} Performance comparison of the base SAM 2.1 model across various prompting methods (MaxVis/MaxSpread) on COCO, OCHuman and CIHP datasets. 6 keypoints proved to be the \textbf{optimal number for prompting}.
%     }
% \label{tab:base-selection-kpts}
% \end{table}

We also briefly explored negative keypoint prompting; however, as it did not yield consistent gains and introduces strong context dependence, we defer a detailed analysis to the Supplementary.

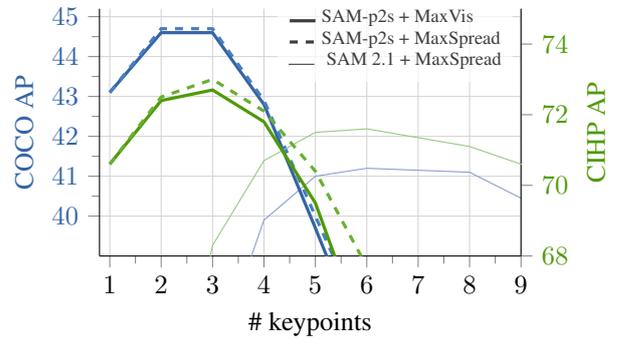
\begin{figure}[tb]
\centering

% Colors (same as previous figure)
\definecolor{coco}{RGB}{52, 101, 164}         % blue
\definecolor{cocoAlt}{RGB}{72, 126, 190} % lighter / more vivid blue
\definecolor{cocoSoft}{RGB}{135, 155, 210} % soft, desaturated blue

\definecolor{cihp}{RGB}{78, 154, 6}        % green
\definecolor{cihpAlt}{RGB}{104, 176, 48} % lighter / more vivid green
\definecolor{cihpSoft}{RGB}{150, 199, 130} % soft, desaturated green

\begin{tikzpicture}
% ---------- LEFT AXIS: COCO ----------
\begin{axis}[
    width=0.88\linewidth,
    height=0.6\linewidth,
    xlabel={\# keypoints},
    ylabel={COCO AP},
    ylabel style={color=coco},
    yticklabel style={color=coco},
    xmin=0.8, xmax=9,
    ymin=39, ymax=45.2,
    xtick={1,2,...,17},
    ytick={40,41,...,50},
    grid=major,
    major grid style={line width=0.3pt, draw=gray!35},
    % minor grid style={line width=0.2pt, draw=gray!20},
    minor tick num=1,
    tick align=outside,
    axis y line*=left,
    axis x line*=bottom,
]

% --- COCO (blue) ---

% SAM 2.1 + MaxSpread
\addplot+[thin, color=cocoSoft, mark=none, mark options={draw=none, fill=none}, mark size=1.6pt]
coordinates {(3,36.2) (4,39.9) (5,41.0) (6,41.2) (8,41.1) (12,38.5) (17,35.7)}; % (2,27.1) 

% SAM-pose2seg + MaxVis solid square
\addplot+[very thick, color=coco, mark=none, mark options={draw=coco, fill=coco}, mark size=1.6pt]
coordinates {(1,43.1) (2,44.6) (3,44.6) (4,42.8) (5,39.7) (6,36.5)};
% SAM-pose2seg + MaxSpread  square
\addplot+[very thick, dashed, color=cocoAlt, mark=none, mark options={draw=cocoAlt, fill=cocoAlt}, mark size=1.6pt]
coordinates {(1,43.1) (2,44.7) (3,44.7) (4,42.9) (5,40.0) (6,37.1)};

\end{axis}

% --- CIHP (green) ---
% ---------- RIGHT AXIS: CIHP ----------
\begin{axis}[
    width=0.88\linewidth,
    height=0.6\linewidth,
    xmin=0.8, xmax=9,
    ymin=68, ymax=75,
    ytick={60,62,...,80},
    ylabel={CIHP AP},
    ylabel style={color=cihp},
    yticklabel style={color=cihp},
    axis y line*=right,
    axis x line=none,
    tick align=outside,
]

% base
% \addplot+[very thick, color=cihp, mark=triangle*, mark options={draw=cihp, fill=cihp}, mark size=1.8pt]
% coordinates {(1,35.8) (2,51.0) (3,60.1)};

\addplot+[thin, color=cihpSoft, mark=none, mark options={draw=none, fill=none}, mark size=1.8pt]
coordinates {(2,61.2) (3,68.3) (4,70.7) (5,71.5) (6,71.6) (8,71.1) (12,69.1) (17,67.9)};

% SA-pose2seg + MaxVis
% \addplot+[very thick, color=cihp, mark=*, mark options={draw=cihp, fill=cihp}, mark size=1.7pt]
\addplot+[very thick, color=cihp, mark=none, mark options={draw=none, fill=none}]
coordinates {(1,70.6) (2,72.4) (3,72.7) (4,71.8) (5,69.5) (6,65.5)};
% SAM-pose2seg + MaxSpread
% \addplot+[very thick, color=cihpAlt, mark=square, mark options={draw=cihpAlt, fill=cihpAlt}, mark size=1.7pt]
\addplot+[very thick, dashed, color=cihpAlt, mark=none, mark options={draw=none, fill=none}]
coordinates {(1,70.6) (2,72.5) (3,73.0) (4,72.1) (5,70.4) (6,67.7)};

\node[
    anchor=north east,
    font=\scriptsize,
    fill=white,
    text=DarkerGray,
    fill opacity=0.7,
    text opacity=1,
    inner sep=4pt
] at (axis description cs:0.98,1.05)
{
\begin{tabular}{@{}l@{}}
\tikz{\draw[very thick] (0,0.5)--(0.5,0.5);} SAM-p2s + MaxVis\\
\tikz{\draw[very thick, dashed] (0,0.5)--(0.5,0.5);} SAM-p2s + MaxSpread\\
\tikz{\draw[thin] (0,0.5)--(0.5,0.5);} \hspace{0.38em}SAM 2.1 + MaxSpread\\
\end{tabular}
};

\end{axis}

\end{tikzpicture}

\caption{
\textbf{Keypoint selection methods for SAM 2.1 and SAM-pose2seg} on \textcolor{coco}{COCO} and \textcolor{cihp}{CIHP} datasets.
SAM 2.1 with MaxSpread (thin line) peaks at 6 keypoints for both datasets.
SAM-pose2seg with MaxSpread (dashed) and MaxVis (full) behaves the same on both datasets and peaks at 3 keypoints.
We selected MaxVis method due to its simplicity.
SAM-pose2seg outperforms SAM 2.1 with both selection methods on both datasets.
}
\label{fig:method_selection_SAM-p2s}

\end{figure}

\subsubsection{Bounding Boxes}
\label{sec:exp-SOTA-segm}

Bounding boxes are not part of our final pipeline, but we analyze their impact as an ablation to better understand SAM’s behavior under pose-guided prompting.

\begin{figure}[tb]
    \centering

    % --- COLUMN 1: LEFT THIRD (Input) ---
    % Trim: Removes the right 2/3 (0.666)
    \begin{subfigure}{0.32\linewidth}
        \centering
        \adjincludegraphics[width=\linewidth, trim={0 0 {0.666\width} 0}, clip]{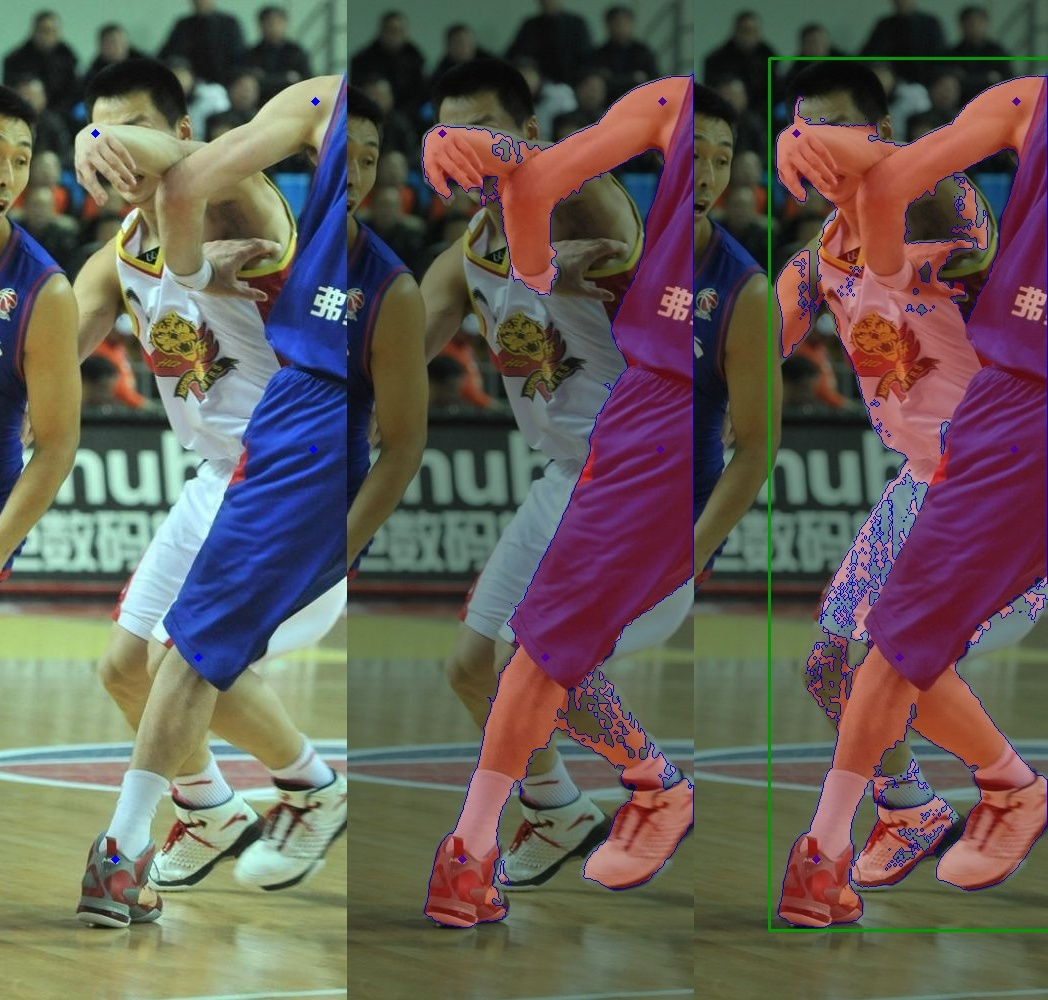}
        \caption*{Input}
    \end{subfigure}
    \hfill
    % --- COLUMN 2: MIDDLE THIRD (Base SAM 2.1) ---
    % Trim: Removes left 1/3 (0.333) and right 1/3 (0.333)
    \begin{subfigure}{0.32\linewidth}
        \centering
        \adjincludegraphics[width=\linewidth, trim={{0.333\width} 0 {0.333\width} 0}, clip]{imgs/SAM-pose2seg/combined_OCHuman005019_1.png}
        \caption*{SAM 2.1}
    \end{subfigure}
    \hfill
    % --- COLUMN 3: RIGHT THIRD (With BBox) ---
    % Trim: Removes left 2/3 (0.666)
    \begin{subfigure}{0.32\linewidth}
        \centering
        \adjincludegraphics[width=\linewidth, trim={{0.666\width} 0 0 0}, clip]{imgs/SAM-pose2seg/combined_OCHuman005019_1.png}
        \caption*{SAM 2.1 + \textcolor{green}{BBox}}
    \end{subfigure}

    \caption{
    \textbf{Problematic bounding box usage} in SAM 2 when multiple people are included. Note: a part of the other person’s hand is recognized incorrectly in both cases due to an incorrect pose keypoint.
    }
    \label{fig:SAM-pose2seg-probl-bbox}
\end{figure}

\begin{figure}[tb]
    \centering

    % --- COLUMN 1: LEFT THIRD (Input) ---
    % Trim: Removes the right 2/3 (0.666)
    \begin{subfigure}{0.32\linewidth}
        \centering
        % Image 1 (Top)
        \adjincludegraphics[width=\linewidth, trim={0 0 {0.666\width} 0}, clip]{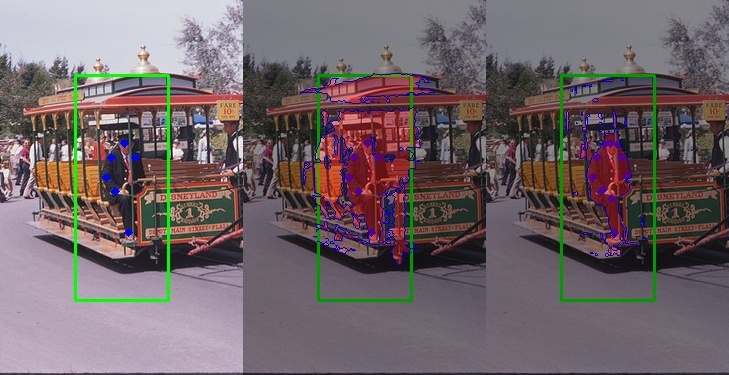}
        \vspace{2pt}
        % Image 2 (Bottom)
        \adjincludegraphics[width=\linewidth, trim={0 0 {0.666\width} 0}, clip]{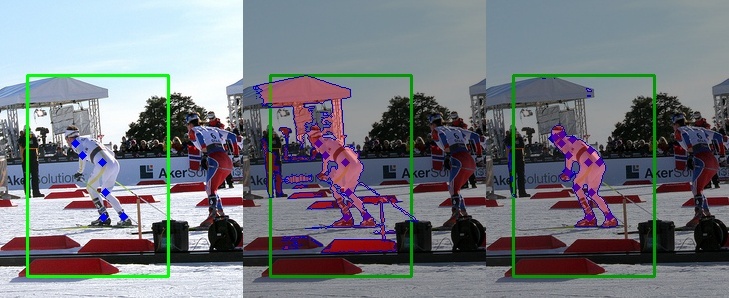}
        \caption*{Input}
    \end{subfigure}
    \hfill
    % --- COLUMN 2: MIDDLE THIRD (Base SAM 2.1) ---
    % Trim: Removes left 1/3 (0.333) and right 1/3 (0.333)
    \begin{subfigure}{0.32\linewidth}
        \centering
        \adjincludegraphics[width=\linewidth, trim={{0.333\width} 0 {0.333\width} 0}, clip]{imgs/SAM-pose2seg/combined_COCO000000016228_0.jpg}
        \vspace{2pt}
        \adjincludegraphics[width=\linewidth, trim={{0.333\width} 0 {0.333\width} 0}, clip]{imgs/SAM-pose2seg/combined_COCO000000438955_1.jpg}
        \caption*{SAM 2.1 + \textcolor{green}{BBox}}
    \end{subfigure}
    \hfill
    % --- COLUMN 3: RIGHT THIRD (SAM 1) ---
    % Trim: Removes left 2/3 (0.666)
    \begin{subfigure}{0.32\linewidth}
        \centering
        \adjincludegraphics[width=\linewidth, trim={{0.666\width} 0 0 0}, clip]{imgs/SAM-pose2seg/combined_COCO000000016228_0.jpg}
        \vspace{2pt}
        \adjincludegraphics[width=\linewidth, trim={{0.666\width} 0 0 0}, clip]{imgs/SAM-pose2seg/combined_COCO000000438955_1.jpg}
        \caption*{SAM 1 + \textcolor{green}{BBox}}
    \end{subfigure}

    \caption{
    \textbf{Inflated bounding box usage} in base SAM 2.1 (middle) and SAM 1 (right). SAM 2.1 often incorporates unnecessary noise when the prompted bounding box does not fit the person exactly. In contrast, SAM 1 performance generally does not drop drastically.
    }
    \label{fig:SAM-pose2seg-probl-noisybox}
\end{figure}

The SAM model does support the use of bounding boxes as prompts, and since ProbPose provides bounding box predictions, it would be natural to incorporate them into our pipeline. Ground-truth bounding boxes were found to be beneficial, as they define precise instance boundaries (see \cref{tab:SAM-pose2seg-bbox}). Even in this case, difficult settings, where a correctly identified bounding box contains large parts of multiple people, might pose a challenge. (see \cref{fig:SAM-pose2seg-probl-bbox}). Nevertheless, bounding boxes predicted by various detectors (that would be a part of iterative pose refinement) may not be as precise \cite{BMPv1}. As a result, segmentation precision degrades significantly when using SAM 2.1: the model may incorrectly include parts of other people or background regions in order to fill the provided bounding box (see \cref{fig:SAM-pose2seg-probl-noisybox}. 

\begin{table}[tb]
\centering
\begin{tabular}{lcccc}
\toprule
Model &
Bbox type &
COCO AP &
CIHP AP \\
\midrule
SAM 2.1 & GT & 50.5 & 76.4 \\
SAM 2.1 & inflated GT & 17.5 & 52.4 \\
SAM 2.1 & none  & 41.2 & 71.6 \\
\midrule
SAM 1   & GT & 52.5 & 72.2 \\
SAM 1 & inflated GT & 45.2 & 59.2 \\
SAM 1   & none & 43.0  & 65.6 \\
\bottomrule
\end{tabular}

\caption{
    \textbf{Usage of bounding boxes in SAM 1 and SAM 2.1.} on COCO val and CIHP val.
    Performance comparison of the base SAM 1 and SAM 2.1 models if ProbPose keypoints are accompanied by bounding boxes.
    We use the keypoint selection method MaxSpread$_6$. While ground-truth bounding boxes seem to be helpful in both scenarios, if both dimensions are enlarged to simulate possible detector noise (by 50 \% in each direction -- 400 \% area increase in total), SAM 2.1 is not reliable.
    }
\label{tab:SAM-pose2seg-bbox}
\end{table}

\subsubsection{Backbone Choice: SAM 1 vs. SAM 2}
\label{sec:SAM 2-SAM 1}
As we tried to analyze whether we could improve our prompting method, we also turned our attention to the older SAM 1 model. In both cases, the second largest models were used (\textit{Hiera Base Plus} for SAM 2 and \textit{Vit-L} for SAM 1). We conducted several tests showing its strengths and weaknesses, and it seems that for our task, SAM 2.1 is more suitable. SAM 1 performs slightly better on COCO dataset, where the poses are more visible, yet lags behind in OCHuman and CIHP datasets. 

It also seems that SAM 1 is able to make better use of negative keypoints in the case of OCHuman and CIHP datasets. When sampling the closest pose keypoints, it serves better as an option to remove superficial parts. Imprecise bounding boxes also do not harm the prediction process significantly, which could not be said for SAM 2.1 that often includes other objects just to fill the boundaries.

However, in general, SAM 2.1 turned out to be a stronger base for our task, as the model itself is lighter (and, therefore, much faster during inference), the provided training code serves well for the fine-tuning and the experiments generally yielded worse results.

\subsubsection{SAM-pose2seg Performance In Challenging Scenes}
\label{sec:SAM-pose2seg-probl-improvements}
\paragraph{\textcolor{blue}{Incomplete segmentation.}}
In contrast to the base SAM 2.1, incomplete segmentation occurs far less frequently. Since SAM-pose2seg is fine-tuned specifically for human segmentation, it learns to recover the full human body more consistently. This addresses a common ambiguity of the base SAM model, which in some cases produced masks corresponding only to clothing, skin regions, or isolated body parts rather than complete human instances.

\paragraph{\textcolor{cyan}{Partially occluded pose.}}
SAM-pose2seg demonstrates precise segmentation in scenarios where pose information is unreliable or heavily occluded. By prioritizing keypoints with the highest visibility scores, the model effectively leverages strong, reliable cues, reducing the impact of missing or ambiguous points. In cases where a single additional misdetected keypoint could compromise the segmentation, SAM-pose2seg maintains accuracy and produces consistent, complete masks.

%%%%%%%%%%%%%%%%%%%%%%%%%%%%%%%%%%%%%%%%%%%%%%%%%%%%%%%%%%%%%%%%%%%%%%%%%%%%%%%%%%%%%%%%%%%%%%%
%%%%%%%%%%%%%%%%%%%%%%%%%%%%%%%%%%%%%%%%%%%%%%%%%%%%%%%%%%%%%%%%%%%%%%%%%%%%%%%%%%%%%%%%%%%%%%%
% ---  ---  ---  ---  ---  ---  ---  ---  --- Intro ---  ---  ---  ---  ---  ---  ---  ---  ---
%%%%%%%%%%%%%%%%%%%%%%%%%%%%%%%%%%%%%%%%%%%%%%%%%%%%%%%%%%%%%%%%%%%%%%%%%%%%%%%%%%%%%%%%%%%%%%%
%%%%%%%%%%%%%%%%%%%%%%%%%%%%%%%%%%%%%%%%%%%%%%%%%%%%%%%%%%%%%%%%%%%%%%%%%%%%%%%%%%%%%%%%%%%%%%%
\section{Conclusions}
\label{sec:conclusions}

To sum up, we achieved substantial improvements in the pose-to-segmentation task for humans by adapting and fine-tuning SAM. We identified and validated the most effective keypoint selection strategies, demonstrating that leveraging \emph{visibility} scores allows the model to reliably choose strong, informative points. Simple and clear prompting methods can be particularly useful in occluded or crowded scenarios.

Our proposed SAM-pose2seg model shows strong generalization and robustness. It excels at recovering full human instances even when most keypoints are missing or unreliable, and is less sensitive to errors that would otherwise harm segmentation in the base SAM model. This makes it suitable both for iterative refinement in pose estimation pipelines and as a standalone tool for human instance segmentation.
The deployment of SAM-pose2seg in the BMP loop \cite{BMPv1} improves the segmentation accuracy from 31.8 AP to 33.7 AP on OCHuman.

Despite these improvements, some limitations remain. SAM-pose2seg can struggle when multiple people are closely overlapping and even high-\emph{visibility} keypoints carry ambiguous semantic meaning, occasionally merging visually similar regions across instances. Future work could explore adaptive keypoint weighting, explicitly modeling the semantic meaning of each keypoint, or integrating the approach into SAM3 to leverage its language-based reasoning capabilities.

Overall, our results demonstrate that SAM-pose2seg provides a practical and reliable approach for human pose-guided segmentation, with potential applications as a pseudo-annotation tool for large-scale datasets or as a robust component in downstream vision pipelines.

% BMP with MaskPose + SAM 2.1     : 31.8/32.2
% BMP with MaskPose + SAM-pose2seg: 33.7/33.6

\textbf{Acknowledgements}.
This work was supported by
% the Technology Agency of the Czech Republic project No. SS05010008, % Picek - rysy
the Ministry of the Interior of the Czech Republic project No. VJ02010041, % FACIS
the Technology Agency of the Czech Republic project CEDMO 2.0 No. FW10010387, % MEDIAN
the European Union’s Digital Europe Programme under Contract No. 101158609, % CEDMO 2.0
and the Czech Technical University student grant SGS23/173/OHK3/3T/13.

%%%%%%%%% REFERENCES
{\small
\bibliographystyle{ieeenat_fullname}
\bibliography{references}

@String(CVPR  = {Proc. CVPR})

@String(ICCV  = {Proc. ICCV})

@String(ECCV  = {Proc. ECCV})

@String(ICPR  = {Proc. ICPR})

@String(AAAI = {Proc. AAAI})

@misc{ttattg,
      title={Test-time Adaptation vs. Training-time Generalization: A Case Study in Human Instance Segmentation using Keypoints Estimation}, 
      author={Kambiz Azarian and Debasmit Das and Hyojin Park and Fatih Porikli},
      year={2022},
      eprint={2212.06242},
      archivePrefix={arXiv},
      primaryClass={cs.CV},
      url={https://arxiv.org/abs/2212.06242}, 
}

@INPROCEEDINGS{pose2body,
  author={Li, Zhong and Chen, Xin and Zhou, Wangyiteng and Zhang, Yingliang and Yu, Jingyi},
  booktitle={2019 IEEE International Conference on Multimedia and Expo (ICME)}, 
  title={Pose2Body: Pose-Guided Human Parts Segmentation}, 
  year={2019},
  volume={},
  number={},
  pages={640-645},
  doi={10.1109/ICME.2019.00116}}

@article{pose2instance,
  title={Pose2Instance: Harnessing Keypoints for Person Instance Segmentation},
  author={Subarna Tripathi and Maxwell D. Collins and Matthew A. Brown and Serge J. Belongie},
  journal={ArXiv},
  year={2017},
  volume={abs/1704.01152},
  url={https://api.semanticscholar.org/CorpusID:13806071}
}

@inproceedings{PosePlusSeg,
  title={Joint Human Pose Estimation and Instance Segmentation with PosePlusSeg},
  author={Niaz Ahmad and Jawad Khan and Jeremy Yuhyun Kim and Youngmoon Lee},
  booktitle={AAAI Conference on Artificial Intelligence},
  year={2022},
  url={https://api.semanticscholar.org/CorpusID:250297175}
}

@article{MultiPoseSeg,
  title={MultiPoseSeg: Feedback Knowledge Transfer for Multi-Person Pose Estimation and Instance Segmentation},
  author={Niaz Ahmad and Jawad Khan and Jeremy Yuhyun Kim and Youngmoon Lee},
  journal={2022 26th International Conference on Pattern Recognition (ICPR)},
  year={2022},
  pages={2086-2092},
  url={https://api.semanticscholar.org/CorpusID:254102117}
}

@article{AIC,
  title={AI Challenger : A Large-scale Dataset for Going Deeper in Image Understanding},
  author={Jiahong Wu and He Zheng and Bo Zhao and Yixin Li and Baoming Yan and Rui Liang and Wenjia Wang and Shipei Zhou and Guosen Lin and Yanwei Fu and Yizhou Wang and Yonggang Wang},
  journal={ArXiv},
  year={2017},
  volume={abs/1711.06475},
  url={https://api.semanticscholar.org/CorpusID:34240712}
}

@article{CrowdHuman,
  title={CrowdHuman: A Benchmark for Detecting Human in a Crowd},
  author={Shuai Shao and Zijian Zhao and Boxun Li and Tete Xiao and Gang Yu and Xiangyu Zhang and Jian Sun},
  journal={ArXiv},
  year={2018},
  volume={abs/1805.00123},
  url={https://api.semanticscholar.org/CorpusID:19933351}
}

@article{RTMDet,
  title={RTMDet: An Empirical Study of Designing Real-Time Object Detectors},
  author={Chengqi Lyu and Wenwei Zhang and Haian Huang and Yue Zhou and Yudong Wang and Yanyi Liu and Shilong Zhang and Kai Chen},
  journal={ArXiv},
  year={2022},
  volume={abs/2212.07784},
  url={https://api.semanticscholar.org/CorpusID:254685870}
}

@article{PoseSegTTA,
  title={Test-time Adaptation vs. Training-time Generalization: A Case Study in Human Instance Segmentation using Keypoints Estimation},
  author={Kambiz Azarian and Debasmit Das and Hyojin Park and Fatih Murat Porikli},
  journal={2023 IEEE/CVF Winter Conference on Applications of Computer Vision Workshops (WACVW)},
  year={2022},
  pages={411-420},
  url={https://api.semanticscholar.org/CorpusID:254591780}
}

@article{ConvNeXt,
  title={A ConvNet for the 2020s},
  author={Liu, Zhuang and Mao, Hanzi and Wu, Chao-Yuan and Feichtenhofer, Christoph and Darrell, Trevor and Xie, Saining},
  journal={Proceedings of the IEEE/CVF Conference on Computer Vision and Pattern Recognition (CVPR)},
  year={2022}
}

@article{SAM2,
  title={Sam 2: Segment anything in images and videos},
  author={Ravi, Nikhila and Gabeur, Valentin and Hu, Yuan-Ting and Hu, Ronghang and Ryali, Chaitanya and Ma, Tengyu and Khedr, Haitham and R{\"a}dle, Roman and Rolland, Chloe and Gustafson, Laura and others},
  journal={arXiv preprint arXiv:2408.00714},
  year={2024}
}

@inproceedings{pose2seg,
  title={Pose2seg: Detection free human instance segmentation},
  author={Zhang, Song-Hai and Li, Ruilong and Dong, Xin and Rosin, Paul and Cai, Zixi and Han, Xi and Yang, Dingcheng and Huang, Haozhi and Hu, Shi-Min},
  booktitle={Proceedings of the IEEE/CVF conference on computer vision and pattern recognition},
  pages={889--898},
  year={2019}
}

@techreport{kmeans++,
  title={k-means++: The advantages of careful seeding},
  author={Arthur, David and Vassilvitskii, Sergei},
  year={2006},
  institution={Stanford}
}

@inproceedings{COCO,
  title={Microsoft COCO: Common Objects in Context},
  author={Tsung-Yi Lin and Michael Maire and Serge J. Belongie and James Hays and Pietro Perona and Deva Ramanan and Piotr Doll{\'a}r and C. Lawrence Zitnick},
  booktitle={European Conference on Computer Vision},
  year={2014}
}

@inproceedings{HRNet,
  title={Deep high-resolution representation learning for human pose estimation},
  author={Sun, Ke and Xiao, Bin and Liu, Dong and Wang, Jingdong},
  booktitle={Proceedings of the IEEE conference on computer vision and pattern recognition},
  pages={5693--5703},
  year={2019}
}

@article{CrowdPose,
  title={CrowdPose: Efficient Crowded Scenes Pose Estimation and A New Benchmark},
  author={Li, Jiefeng and Wang, Can and Zhu, Hao and Mao, Yihuan and Fang, Hao-Shu and Lu, Cewu},
  journal={arXiv preprint arXiv:1812.00324},
  year={2018}
}

@inproceedings{MPII,
               author = {Mykhaylo Andriluka and Leonid Pishchulin and Peter Gehler and Schiele, Bernt},
               title = {2D Human Pose Estimation: New Benchmark and State of the Art Analysis},
               booktitle = {IEEE Conference on Computer Vision and Pattern Recognition (CVPR)},
               year = {2014},
               month = {June}
}

@article{YOLOX,
  title={Yolox: Exceeding yolo series in 2021},
  author={Ge, Z},
  journal={arXiv preprint arXiv:2107.08430},
  year={2021}
}

@inproceedings{coDETR,
  title={Detrs with collaborative hybrid assignments training},
  author={Zong, Zhuofan and Song, Guanglu and Liu, Yu},
  booktitle={Proceedings of the IEEE/CVF international conference on computer vision},
  pages={6748--6758},
  year={2023}
}

@inproceedings{CrowdSAM,
  title={Crowd-SAM: SAM as a Smart Annotator for Object Detection in Crowded Scenes},
  author={Cai, Zhi and Gao, Yingjie and Zheng, Yaoyan and Zhou, Nan and Huang, Di},
  booktitle={Proceedings of the European Conference on Computer Vision (ECCV)},
  year={2024}
}

@ARTICLE{PoSeg,
  author={Zhou, Desen and He, Qian},
  journal={IEEE Access}, 
  title={PoSeg: Pose-Aware Refinement Network for Human Instance Segmentation}, 
  year={2020},
  volume={8},
  number={},
  pages={15007-15016},
  doi={10.1109/ACCESS.2020.2967147}}

@inproceedings{IterDet,
  title={Iterdet: iterative scheme for object detection in crowded environments},
  author={Rukhovich, Danila and Sofiiuk, Konstantin and Galeev, Danil and Barinova, Olga and Konushin, Anton},
  booktitle={Structural, syntactic, and statistical pattern recognition: Joint IAPR international workshops, s+ SSPR 2020, padua, Italy, January 21--22, 2021, proceedings},
  pages={344--354},
  year={2021},
  organization={Springer}
}

@inproceedings{CIHP,
  title={Instance-level human parsing via part grouping network},
  author={Gong, Ke and Liang, Xiaodan and Li, Yicheng and Chen, Yimin and Yang, Ming and Lin, Liang},
  booktitle={Proceedings of the European conference on computer vision (ECCV)},
  pages={770--785},
  year={2018}
}

@InProceedings{BMPv1,
            author    = {Purkrabek, Miroslav and Matas, Jiri},
            title     = {Detection, Pose Estimation and Segmentation for Multiple Bodies: Closing the Virtuous Circle},
            booktitle = {Proceedings of the IEEE/CVF International Conference on Computer Vision (ICCV)},
            month     = {October},
            year      = {2025},
            pages     = {9004-9013}
          }

@InProceedings{ProbPose,
    author    = {Purkrabek, Miroslav and Matas, Jiri},
    title     = {{ProbPose}: A Probabilistic Approach to 2D Human Pose Estimation},
    booktitle = {Proceedings of the Computer Vision and Pattern Recognition Conference (CVPR)},
    month     = {June},
    year      = {2025},
    pages     = {27124-27133}
}

@article{YOLOv3,
  title={Yolov3: An incremental improvement},
  author={Redmon, Joseph and Farhadi, Ali},
  journal={arXiv preprint arXiv:1804.02767},
  year={2018}
}

@inproceedings{ViTDet,
  title={Exploring plain vision transformer backbones for object detection},
  author={Li, Yanghao and Mao, Hanzi and Girshick, Ross and He, Kaiming},
  booktitle={European conference on computer vision},
  pages={280--296},
  year={2022},
  organization={Springer}
}

@inproceedings{SAM1,
  title={Segment anything},
  author={Kirillov, Alexander and Mintun, Eric and Ravi, Nikhila and Mao, Hanzi and Rolland, Chloe and Gustafson, Laura and Xiao, Tete and Whitehead, Spencer and Berg, Alexander C and Lo, Wan-Yen and others},
  booktitle={Proceedings of the IEEE/CVF international conference on computer vision},
  pages={4015--4026},
  year={2023}
}

@article{GroundedSAM,
  title={Grounded sam: Assembling open-world models for diverse visual tasks},
  author={Ren, Tianhe and Liu, Shilong and Zeng, Ailing and Lin, Jing and Li, Kunchang and Cao, He and Chen, Jiayu and Huang, Xinyu and Chen, Yukang and Yan, Feng and others},
  journal={arXiv preprint arXiv:2401.14159},
  year={2024}
}

@inproceedings{SAMWise,
  title={Samwise: Infusing wisdom in sam2 for text-driven video segmentation},
  author={Cuttano, Claudia and Trivigno, Gabriele and Rosi, Gabriele and Masone, Carlo and Averta, Giuseppe},
  booktitle={Proceedings of the Computer Vision and Pattern Recognition Conference},
  pages={3395--3405},
  year={2025}
}

@InProceedings{SA-SAM,
    author    = {Wei, Zhaoyang and Chen, Pengfei and Yu, Xuehui and Li, Guorong and Jiao, Jianbin and Han, Zhenjun},
    title     = {Semantic-aware SAM for Point-Prompted Instance Segmentation},
    booktitle = {Proceedings of the IEEE/CVF Conference on Computer Vision and Pattern Recognition (CVPR)},
    month     = {June},
    year      = {2024},
    pages     = {3585-3594}
}

@article{OpenWorldSAM,
  title={OpenWorldSAM: Extending SAM2 for Universal Image Segmentation with Language Prompts},
  author={Xiao, Shiting and Kabra, Rishabh and Li, Yuhang and Lee, Donghyun and Carreira, Joao and Panda, Priyadarshini},
  journal={arXiv preprint arXiv:2507.05427},
  year={2025}
}

@inproceedings{ParsingRCNN,
  title = {Parsing R-CNN for Instance-Level Human Analysis},
  author = {Lu Yang and Qing Song and Zhihui Wang and Ming Jiang},
  booktitle = {Proceedings of the IEEE Conference on Computer Vision and Pattern Recognition (CVPR)},
  year = {2019}
}

@article{DINOv2,
  title={Dinov2: Learning robust visual features without supervision},
  author={Oquab, Maxime and Darcet, Timoth{\'e}e and Moutakanni, Th{\'e}o and Vo, Huy and Szafraniec, Marc and Khalidov, Vasil and Fernandez, Pierre and Haziza, Daniel and Massa, Francisco and El-Nouby, Alaaeldin and others},
  journal={arXiv preprint arXiv:2304.07193},
  year={2023}
}
}

%%%%%%%%%%%% SUPPL
% \clearpage
\appendix
\maketitlesupplementary
\section{Prompting Experiments}
\subsection{Negative Keypoint Prompting}

While the negative keypoints are conceptually appealing for resolving overlaps, negative keypoints proved difficult to apply effectively. A key challenge is simulating the human corrective behavior intended by the SAM framework when selecting negative prompts.

We evaluated two negative keypoint selection strategies: \textcolor{TealBlue}{(1)} the spatially closest keypoint from a different pose, and \textcolor{violet}{(2)} the target pose keypoint with the lowest \emph{visibility} score. Neither yielded consistent improvements due to high context dependence.

The \textcolor{TealBlue}{first} strategy fails to verify if the sampled keypoint lies within an erroneous prediction region, misaligning with negative prompting's intended use. It also ignores potential occlusions where segmentation should legitimately include the region; excluding occluded points is non-trivial as no fixed \emph{visibility} threshold reliably separates visible from occluded joints.

In the \textcolor{violet}{second} case, the lowest \emph{visibility} score may still indicate a valid body part, as occlusion counts vary. Consequently, visibility scores serve as relative confidence measures rather than binary valid/invalid distinctions, limiting their effectiveness for negative prompting.

\begin{table}[tb]
\centering
\setlength{\tabcolsep}{4pt} % reduce column spacing
\begin{tabular}{lcccc}
\toprule
Model &
\makecell{Neg. kpts\\selection} &
\makecell{COCO\\val AP} &
\makecell{OCHuman\\test AP} &
\makecell{CIHP\\val AP} \\
\midrule
\makecell{SAM 2.1} & \textcolor{TealBlue}{1 least vis.} & 37.5 & 26.4 & 66.9 \\
\makecell{SAM 2.1} & \textcolor{violet}{1 closest} & 39.4 & 27.7 & 69.0 \\
\makecell{SAM 2.1} & none & 41.2 & 29.5 & 71.6 \\
\midrule
\makecell{SAM-pose2seg} & \textcolor{TealBlue}{1 least vis.} & 41.2 & 26.3 & 68.8  \\
\makecell{SAM-pose2seg} & \textcolor{violet}{1 closest} & 40.4 & 35.2 & 70.1 \\
\makecell{SAM-pose2seg} & none & 43.7 & 34.1 & 71.7 \\
\midrule
\makecell{SAM 1} & \textcolor{TealBlue}{1 least vis.} & 40.2 & 28.1 & 63.2 \\
\makecell{SAM 1} & \textcolor{violet}{1 closest} & 40.9 & 28.8 & 66.9 \\
\makecell{SAM 1} & none & 43.0 & 28.1 & 65.6 \\
\bottomrule
\end{tabular}
\caption{
    \textbf{ Usage of negative keypoints in base and fine-tuned model.} Performance comparison of various methods of negative keypoint prompting across the base SAM2 model, SAM-pose2seg and the base SAM 1 model. MaxSpread$_6$ keypoint selection method for base SAM 2.1 and SAM 1, MaxVis$_3$ for SAM-pose2seg. Unlike in SAM 1, negative keypoints seemingly do not appear to improve segmentation quality in SAM 2.1.
    }
\label{tab:base-vs-finetuned-selection-kpts}
\end{table}

\begin{figure}[tb]
    \centering

    % --- COLUMN 1: LEFT THIRD (Input) ---
    % Trim: Removes the right 2/3 (0.666)
    \begin{subfigure}{0.32\linewidth}
        \centering
        % Top Image
        \adjincludegraphics[height=6 cm, width=\linewidth, trim={0 0 {0.7\width} 0}, clip]{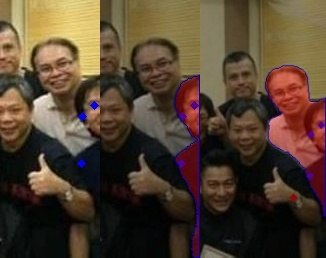}
        \vspace{2pt}
        % Bottom Image
        \adjincludegraphics[width=\linewidth, trim={0 0 {0.666\width} 0}, clip]{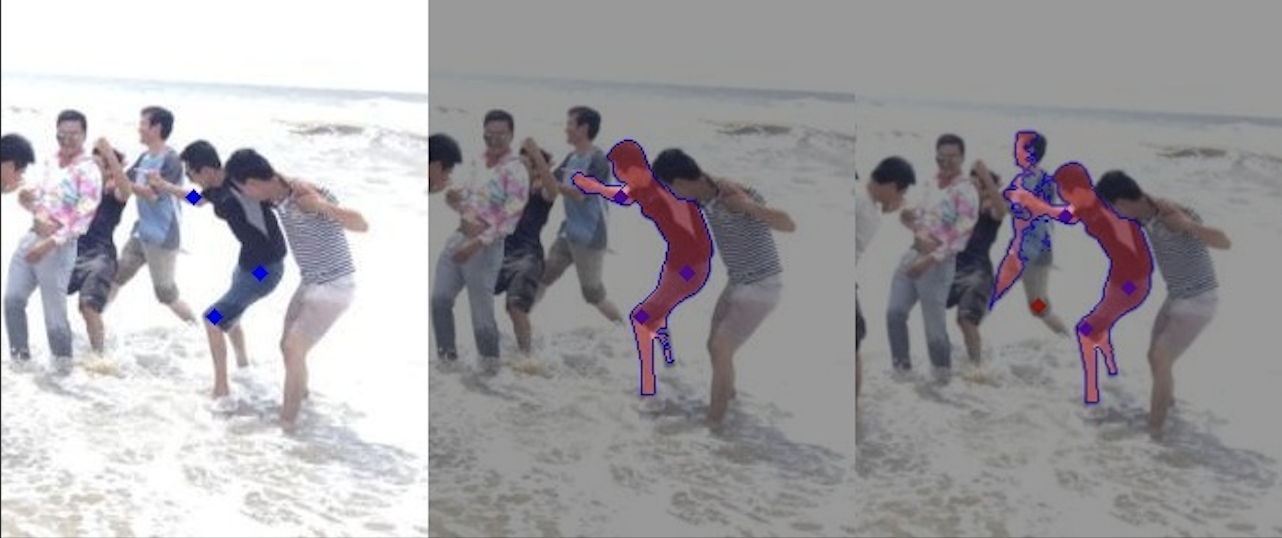}
        \caption*{Input}
    \end{subfigure}
    \hfill
    % --- COLUMN 2: MIDDLE THIRD (Without Negative Keypoint) ---
    % Trim: Removes left 1/3 (0.333) and right 1/3 (0.333)
    \begin{subfigure}{0.32\linewidth}
        \centering
        \adjincludegraphics[height=6 cm, width=\linewidth, trim={{0.3\width} 0 {0.4\width} 0}, clip]{imgs/SAM-pose2seg/combined_CIHP0033255_11.png}
        \vspace{2pt}
        \adjincludegraphics[width=\linewidth, trim={{0.333\width} 0 {0.333\width} 0}, clip]{imgs/SAM-pose2seg/combined_CIHP0037748_7.png}
        \caption*{Without Neg. Kpt}
    \end{subfigure}
    \hfill
    % --- COLUMN 3: RIGHT THIRD (With Negative Keypoint) ---
    % Trim: Removes left 2/3 (0.666)
    \begin{subfigure}{0.32\linewidth}
        \centering
        \adjincludegraphics[height=4.8 cm, width=\linewidth, trim={{0.6\width} 0 0 0}, clip]{imgs/SAM-pose2seg/combined_CIHP0033255_11.png}
        \vspace{2pt}
        \adjincludegraphics[width=\linewidth, trim={{0.666\width} 0 0 0}, clip]{imgs/SAM-pose2seg/combined_CIHP0037748_7.png}
        \caption*{With Neg. Kpt}
    \end{subfigure}

    \caption{
    \textbf{Drawbacks of negative keypoint usage} - picking a negative keypoint can cause an unpredictable shift in the segmentation. Comparison of SAM-pose2seg without and with a closest negative keypoint (colored \textcolor{red}{red}) from another pose.
    }
    \label{fig:SAM-pose2seg-probl-negclose}
\end{figure}

\subsection{Keypoint Occlusion}

During keypoint selection, the primary challenge is avoiding the selection of keypoints belonging to a person whose body parts are occluded by another individual. Leveraging ProbPose, we make use of keypoint-specific confidence measures to reduce the likelihood of choosing unreliable keypoints. In our implementation, we use the keypoint \emph{visibility} parameter, which provides a confidence score in the range [0,1]. Alternative measures, such as \emph{presence probability} and \emph{expected OKS}, were also considered; however, empirical evaluation showed that both were slightly less effective (see \cref{tab:ProbPose-params-kpts}). 
\begin{table}[tb]
\centering

\begin{tabular}{lcccc}
\toprule
\makecell{Model} & \makecell{Method} &
\makecell{COCO\\val AP} &
\makecell{OCHuman\\test AP} &
\makecell{CIHP\\val AP}\\
\midrule
base & vis. & \textbf{41.2} &\textbf{29.5} & \textbf{71.6} \\
base & pres. prob. & 38.2 & 23.0 & 61.8 \\
base & exp. OKS  & 35.4 & 25.3 & 64.5 \\
\midrule
p2s & vis. & \textbf{44.6} & \textbf{34.7} & \textbf{72.7} \\
p2s & pres. prob. & 43.8 & 31.0 & 69.3 \\
p2s & exp. OKS & 43.9 & \textbf{34.7} & 72.0 \\
\bottomrule
\end{tabular}
\caption{
\textbf{Comparison of ProbPose keypoint selection parameters} -- \emph{visibility}, \emph{presence probability} and \emph{expected OKS}.
The first section reports results for the base model (MaxSpread$_6$), where keypoints are initially filtered using a confidence score. Among the three evaluated confidence measures, the \emph{visibility} parameter yields the best overall performance.
The same trend is observed for the SAM-pose2seg (p2s) model. Using method MaxVis$_3$, which ranks keypoints by their score and selects the top three, \emph{visibility} again outperforms the alternative measures.
}
\label{tab:ProbPose-params-kpts}
\end{table}

\begin{figure}[tb]
        \centering
        \begin{subfigure}{0.48\linewidth}
        \centering
        \adjincludegraphics[height=3.4 cm, width=\linewidth, trim={0 0 0 0}, clip]{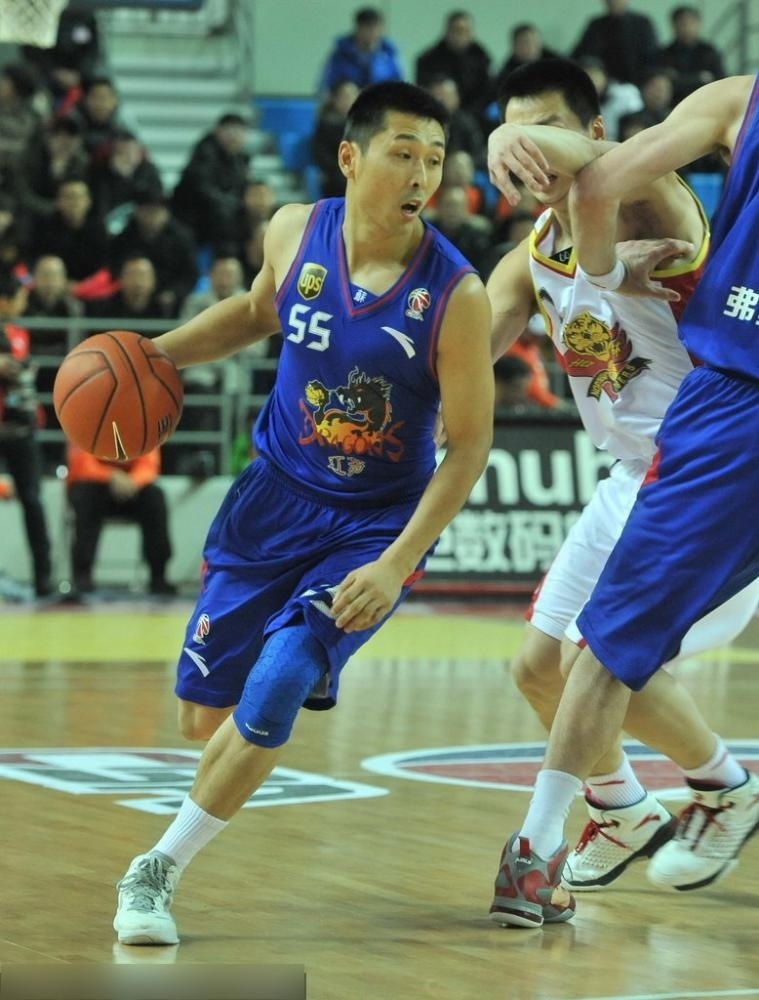}
    \end{subfigure}
    \hfill
    \begin{subfigure}{0.24\linewidth}
        \centering
        \adjincludegraphics[height=3.4 cm, width=\linewidth, trim={0 0 0 0}, clip]{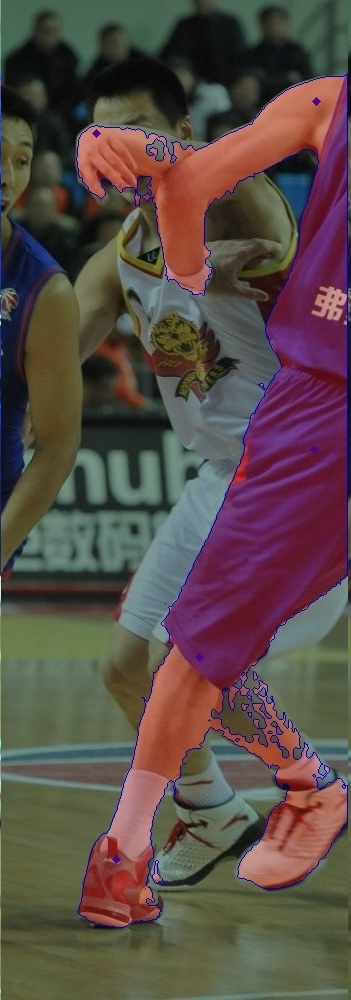}
    \end{subfigure}
    \hfill
    \begin{subfigure}{0.24\linewidth}
        \centering
        \adjincludegraphics[height=3.4 cm, width=\linewidth, trim={0 0 0 0}, clip]{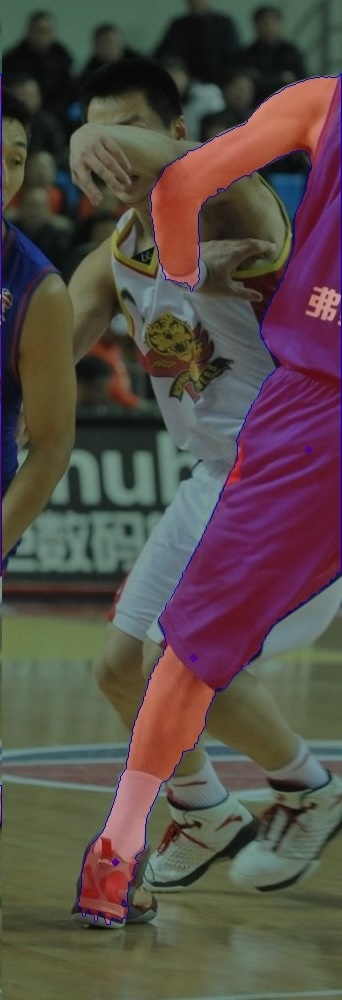}
    \end{subfigure}
    
    \centering
        \begin{subfigure}{0.45\linewidth}
        \centering
        \adjincludegraphics[height=4.8 cm, width=\linewidth, trim={0 0 {0.51\width} 0}, clip]{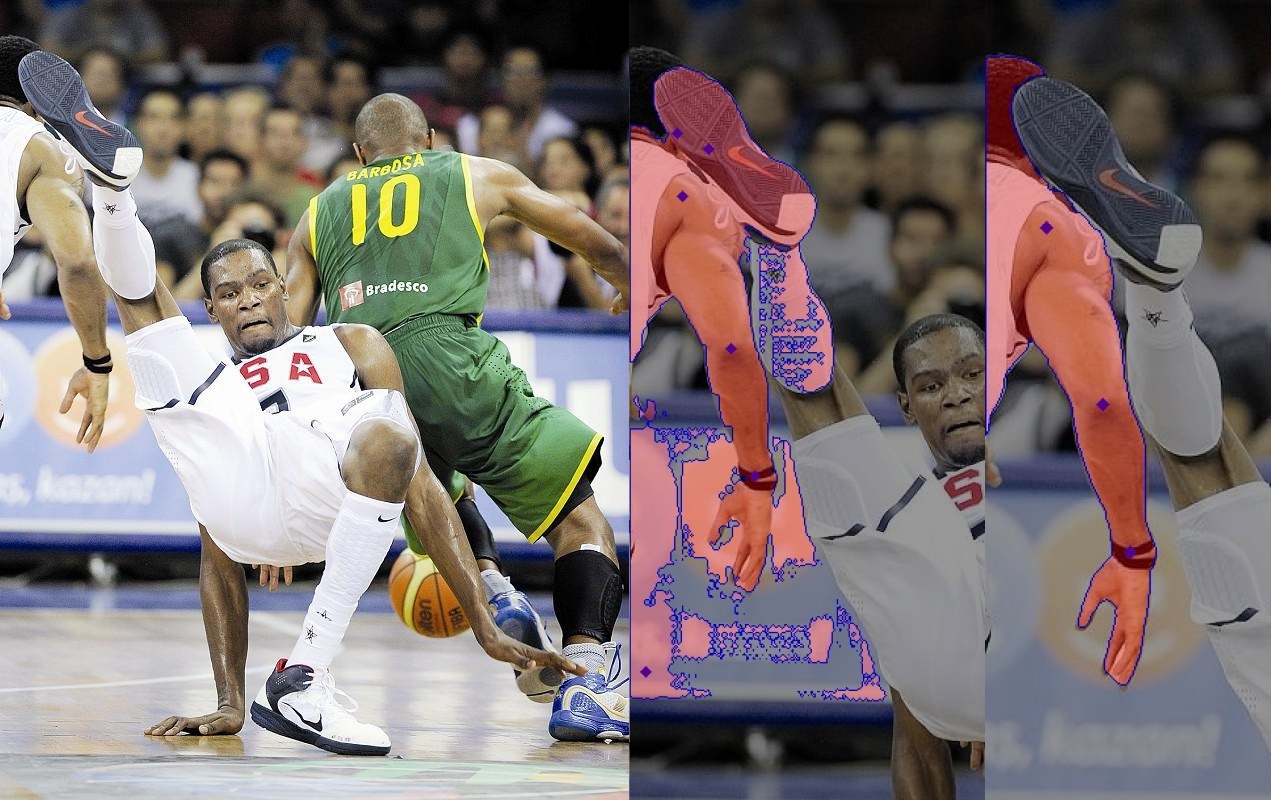}
    \end{subfigure}
    \hfill
    \begin{subfigure}{0.27\linewidth}
        \centering
        \adjincludegraphics[height=4.8 cm, width=\linewidth, trim={{0.5\width} 0 {0.225\width} 0}, clip]{imgs/SAM-pose2seg/combined_OCHuman000903_3.jpg}
    \end{subfigure}
    \hfill
    \begin{subfigure}{0.24\linewidth}
        \centering
        \adjincludegraphics[height=4.8 cm, width=\linewidth, trim={{0.775\width} 0 0 0}, clip]{imgs/SAM-pose2seg/combined_OCHuman000903_3.jpg}
    \end{subfigure}
        
    \centering
    
    \begin{subfigure}{0.32\linewidth}
        \centering
        \adjincludegraphics[height=3.0 cm, width=\linewidth, trim={0 0 {0.7\width} 0}, clip]{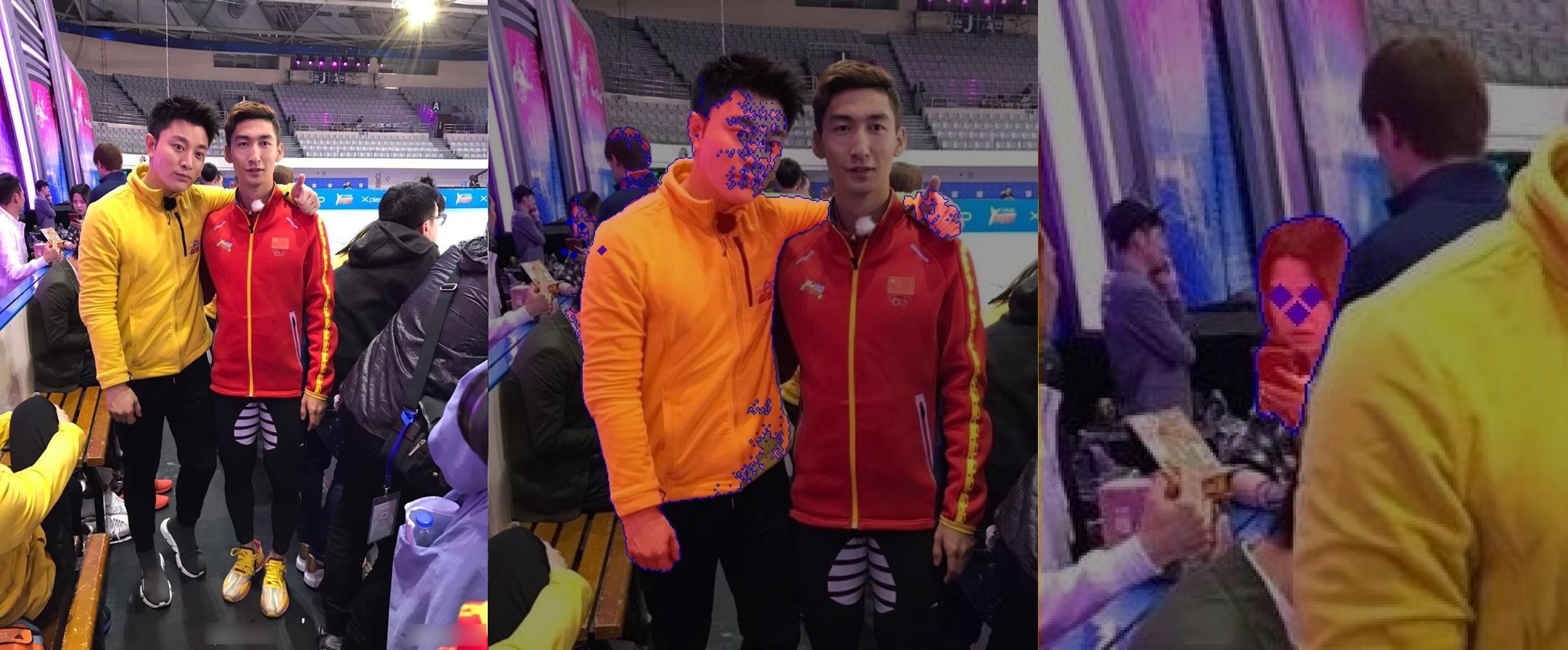}
    \end{subfigure}
    \hfill
    \begin{subfigure}{0.32\linewidth}
        \centering
        \adjincludegraphics[height=3.0 cm, width=\linewidth, trim={{0.334\width} 0 {0.34\width} 0}, clip]{imgs/SAM-pose2seg/combined_OCHuman003445_13.jpg}
    \end{subfigure}
    \hfill
    \begin{subfigure}{0.32\linewidth}
        \centering
        \adjincludegraphics[height=3.0 cm, width=\linewidth, trim={{0.666\width} 0 0 0}, clip]{imgs/SAM-pose2seg/combined_OCHuman003445_13.jpg}
    \end{subfigure}
    
    \begin{subfigure}{0.28\linewidth}
        \centering
        \adjincludegraphics[height=3.7 cm, width=\linewidth, trim={{0.15\width} 0 {0.15\width} 0}, clip]{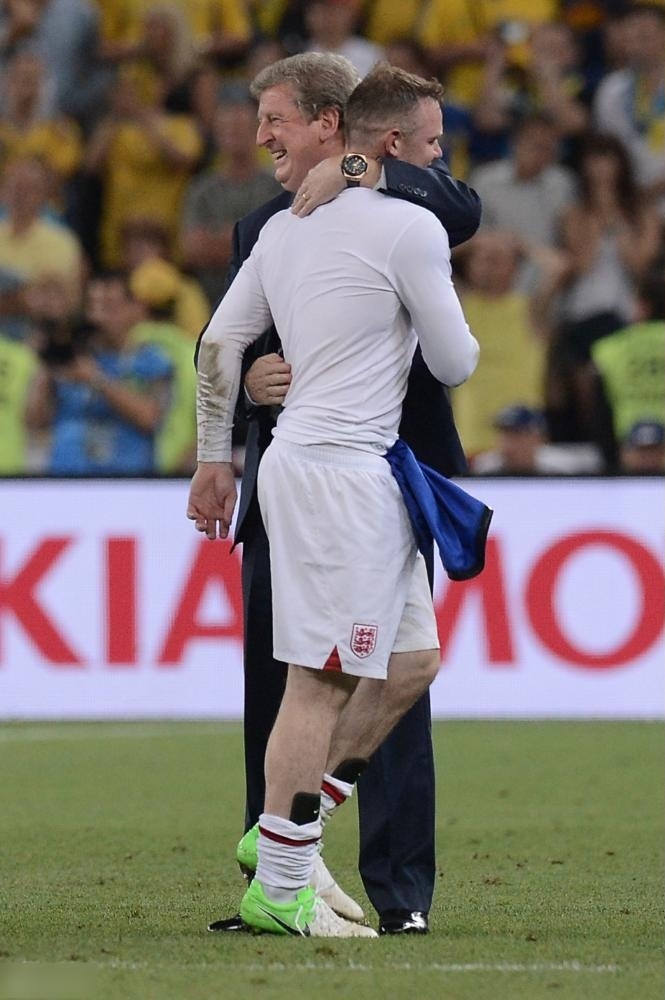}
    \end{subfigure}
    \hfill
    \begin{subfigure}{0.4\linewidth}
        \centering
        \adjincludegraphics[height=3.7 cm, width=\linewidth, trim={0 0 0 0}, clip]{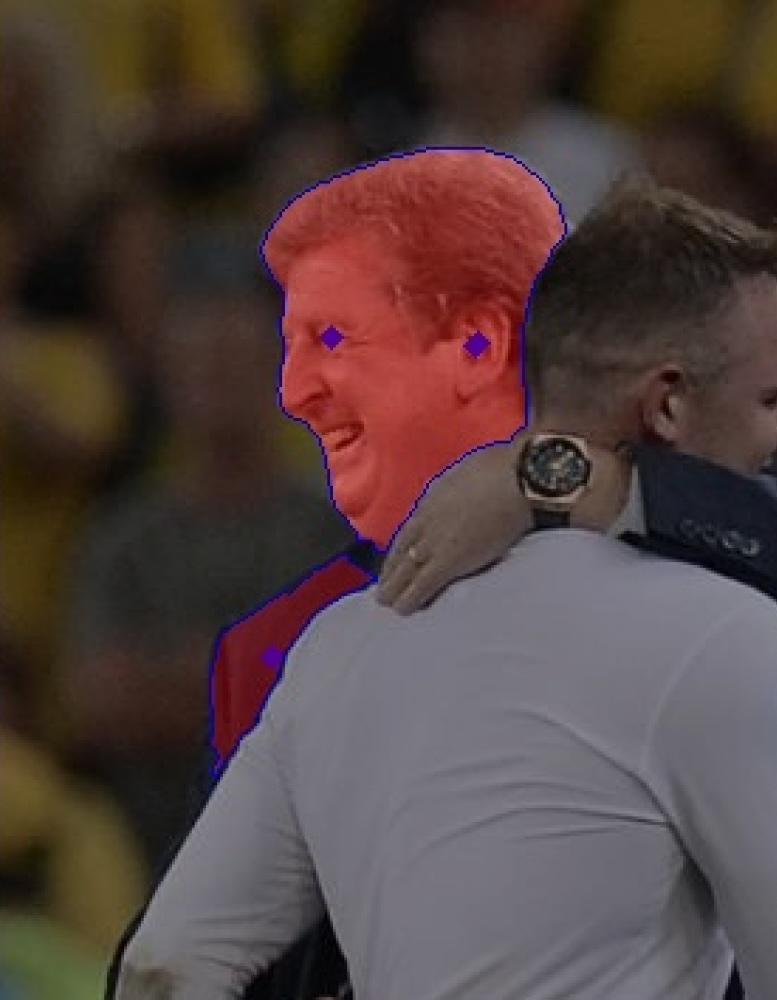}
    \end{subfigure}
    \hfill
    \begin{subfigure}{0.28\linewidth}
        \centering
        \adjincludegraphics[height=3.7 cm, width=\linewidth, trim={0 0 0 0}, clip]{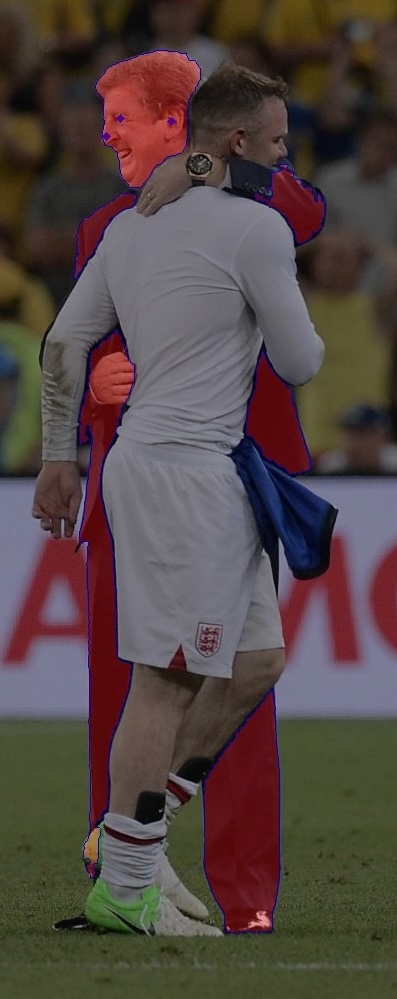}
    \end{subfigure}
    
    \begin{subfigure}{0.38\linewidth}
        \centering
        \adjincludegraphics[height=2.3 cm, width=\linewidth, trim={0 0 {0.6\width} 0}, clip]{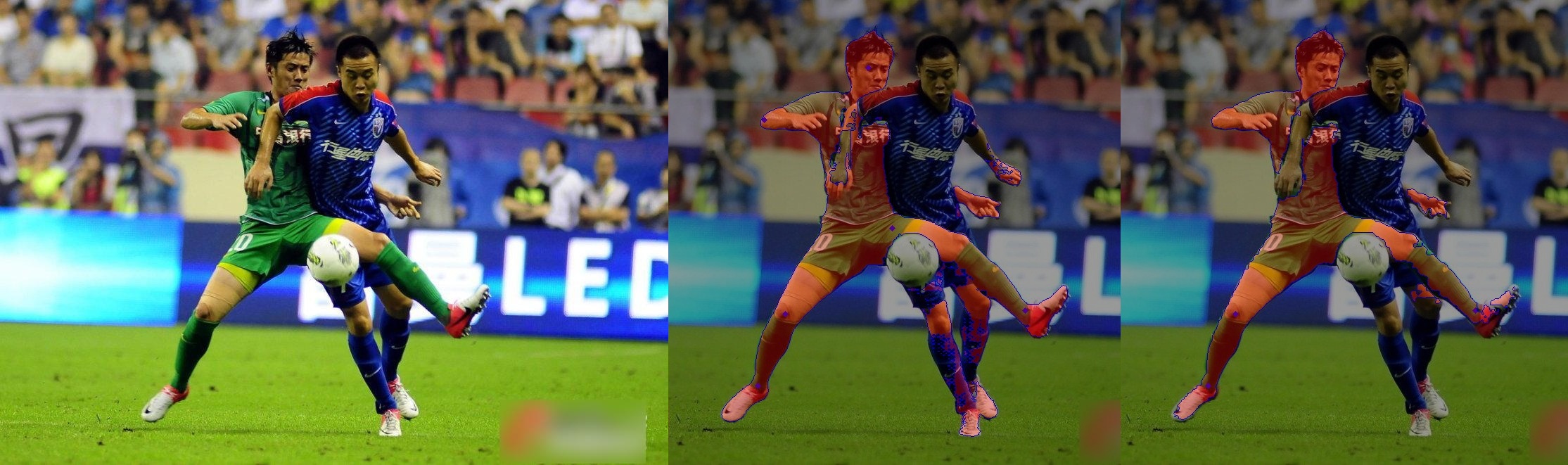}
        \caption*{Input}
    \end{subfigure}
    \hfill
    \begin{subfigure}{0.29\linewidth}
        \centering
        \adjincludegraphics[height=2.3 cm, width=\linewidth, trim={{0.43\width} 0 {0.3\width} 0}, clip]{imgs/SAM-pose2seg/combined_OCHuman002763_0.jpg}
        \caption*{SAM 2.1}
    \end{subfigure}
    \hfill
    \begin{subfigure}{0.29\linewidth}
        \centering
        \adjincludegraphics[height=2.3 cm, width=\linewidth, trim={{0.72\width} 0 0 0}, clip]{imgs/SAM-pose2seg/combined_OCHuman002763_0.jpg}
        \caption*{SAM-pose2seg}
    \end{subfigure}
            
    \caption{
        \textbf{Additional qualitative improvements of SAM-pose2seg over the base SAM 2.1.}
        The first three triplets demonstrate improvements achieved through simple pose-based keypoint selection, while the last two highlight the model’s enhanced pose awareness in more challenging scenarios.  
    \vspace{-2em}
    }
    \label{fig:SAM-pose2seg-probl-improvsuppl}
\end{figure}

\subsection{Image Cropping}
\begin{figure}[tb]
    \centering

    % --- COLUMN 1: LEFT THIRD (Input) ---
    % Trim: Removes the right 2/3 (0.666)
    \begin{subfigure}{0.32\linewidth}
        \centering
        \adjincludegraphics[width=\linewidth, trim={0 0 {0.666\width} 0}, clip]{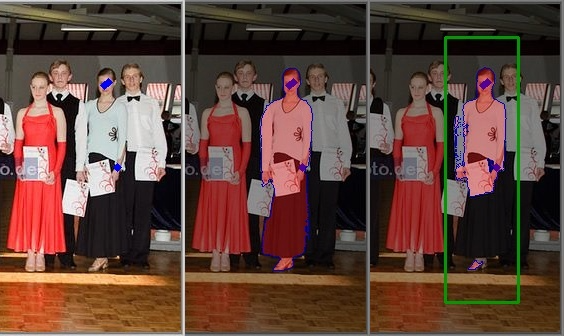}
        \caption*{Input}
    \end{subfigure}
    \hfill
    % --- COLUMN 2: MIDDLE THIRD (Without Crop) ---
    % Trim: Removes left 1/3 (0.333) and right 1/3 (0.333)
    \begin{subfigure}{0.32\linewidth}
        \centering
        \adjincludegraphics[width=\linewidth, trim={{0.333\width} 0 {0.333\width} 0}, clip]{imgs/SAM-pose2seg/combined_CIHP0005526_3.png}
        \caption*{Without Cropping}
    \end{subfigure}
    \hfill
    % --- COLUMN 3: RIGHT THIRD (With Crop) ---
    % Trim: Removes left 2/3 (0.666)
    \begin{subfigure}{0.32\linewidth}
        \centering
        \adjincludegraphics[width=\linewidth, trim={{0.666\width} 0 0 0}, clip]{imgs/SAM-pose2seg/combined_CIHP0005526_3.png}
        \caption*{With Cropping}
    \end{subfigure}

    \caption{
    \textbf{Crop usage} leads to a noisy and incomplete prediction of a human in the foreground. Comparison of base SAM 2.1 without and with image cropping - the \textcolor{green}{green area} is the cropped area.
    }
    \label{fig:SAM-pose2seg-probl-crop}
\end{figure}

Apart from incorporating bounding boxes into our prompting pipleine, we evaluated an approach that performs inference only on cropped image regions instead of the full image, using bounding boxes to reduce noise and instance overlap. However, this strategy did not yield improvements, potentially because SAM is designed to operate on high-resolution images containing a wide variety of objects, and restricting the context may negatively impact its performance (see Image~\cref{fig:SAM-pose2seg-probl-crop}). This method is also significantly time-consuming, as we set an image before every prediction instead of one set per all instances in one image.

\begin{table}[tb]
\centering
\begin{tabular}{lccc}
\toprule
Model & 
Crop type & 
\makecell{COCO\\val} &
\makecell{CIHP\\val} \\
\midrule
SAM 2.1 & 1.5x BBox & 36.8 & 61.1 \\
SAM 2.1 & no crop & 41.2 & 71.6 \\
\midrule
SAM-pose2seg & 1.5x BBox & 34.6 & 60.3 \\
SAM-pose2seg & no crop & 44.6 & 72.7 \\
\bottomrule
\end{tabular}
\caption{
    \textbf{Cropping images to the bounding boxes of instances.} Performance comparison of the base SAM 2.1 and SAM-pose2seg models if ProbPose keypoints are cropped to only preserve the area inside of a ground-truth bounding box (to ensure nothing is left out, we tried inflating bounding box 1.5 times). Cropping did not provide any additional value.
    \vspace{-0.5em}
    }
\label{tab:SAM-pose2seg-cropbox}
\end{table}

\section{Incorrect SAM-pose2seg Segmentation}
Although SAM-pose2seg is trained exclusively on human data, it is not explicitly supervised to learn the semantic meaning of individual pose keypoints. Consequently, the model does not always reliably distinguish between visually similar body parts belonging to different individuals, particularly in crowded or heavily occluded scenes.

\subsection{Visual Similarity}
In some cases, appearance cues such as clothing texture or color dominate the prediction, causing spatially distant regions with similar visual properties to be merged into a single instance. This issue persists across datasets, likely due to the strong influence of the image encoder, which remained frozen to preserve SAM’s generalization capability and to prevent overfitting to human-specific appearance patterns. As a result, visually similar regions may still be favored over pose-based cues in ambiguous scenarios, which remains a known limitation of the approach. Potential improvements mainly lie in more careful keypoint selection, as avoiding occluded or unreliable keypoints reduces the likelihood of producing corrupted segmentation masks (see \cref{fig:SAM-pose2seg-probl-improvsuppl}).

\subsection{Keypoint Information}
In other scenarios, background individuals remain unsegmented due to an insufficient number of reliable keypoints that would provide adequate pose information. Moreover, even keypoints with relatively high \emph{visibility} scores may correspond to occluded body parts; in such cases, incorrect segmentation becomes difficult to avoid.

\begin{figure}[tb]
    \centering

    \begin{subfigure}{0.48\linewidth}
        \centering
        \adjincludegraphics[width=\linewidth, trim={0 0 {.5\width} 0}, clip]{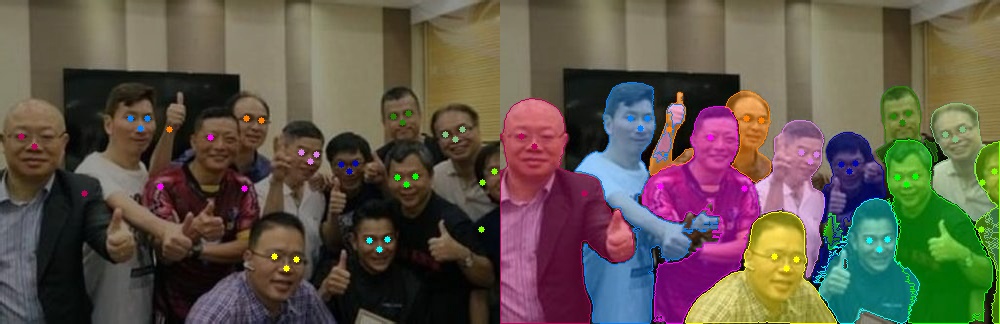}
        \vspace{2pt}
        \adjincludegraphics[width=\linewidth, trim={0 0 {.5\width} 0}, clip]{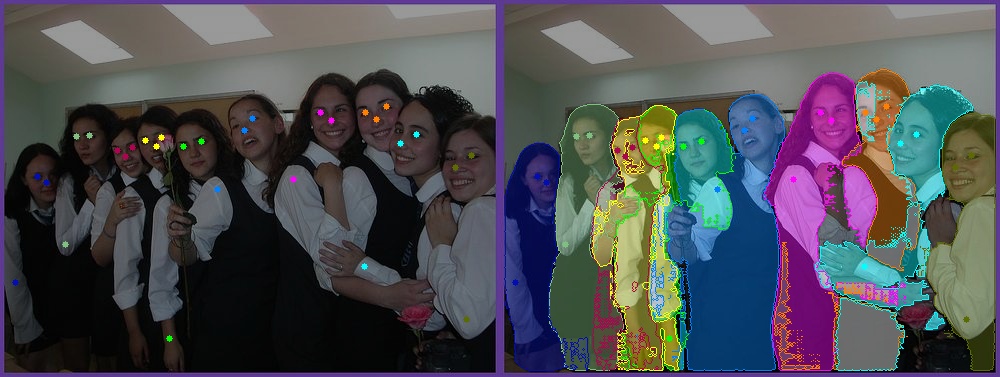}
        \vspace{2pt}
        \adjincludegraphics[width=\linewidth, trim={0 0 0 0}, clip]{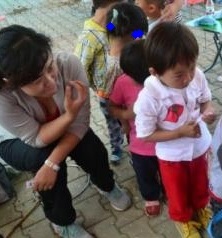}
        
        \caption*{Input}
    \end{subfigure}
    \hfill
    \begin{subfigure}{0.48\linewidth}
        \centering
        \adjincludegraphics[width=\linewidth, trim={{.5\width} 0 0 0}, clip]{imgs/SAM-pose2seg/0033255.jpg}
        \vspace{2pt}
        \adjincludegraphics[width=\linewidth, trim={{.5\width} 0 0 0}, clip]{imgs/SAM-pose2seg/0005045.jpg}
        \vspace{2pt}
        \adjincludegraphics[width=\linewidth, trim={0 0 0 0}, clip]{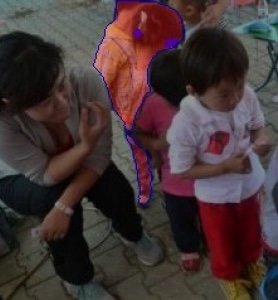}
        \caption*{SAM-pose2seg}
    \end{subfigure}

    \caption{
        \textbf{Persisting overlaps in occluded scenarios} observed in SAM-pose2seg predictions.
    \vspace{17em}
    }
    \label{fig:SAM-pose2seg-probl-overlaps}
\end{figure}

\begin{figure}[tb]
    \centering

    \begin{subfigure}{0.48\linewidth}
        \centering
        \adjincludegraphics[width=\linewidth, trim={0 0 0 0}, clip]{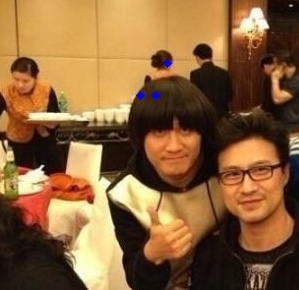}
        \caption*{SAM-pose2seg}
    \end{subfigure}
    \hfill
    \begin{subfigure}{0.48\linewidth}
        \centering
        \adjincludegraphics[width=\linewidth, trim={0 0 0 0}, clip]{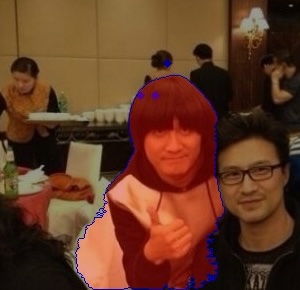}
        \caption*{SAM-pose2seg}
    \end{subfigure}

    \caption{
        \textbf{Visibility score instabilities.} Two out of the three most visible keypoints with a \emph{visibility} score above 0.5 pointed to the occluded limbs.
        \vspace{60em}
    }
    \label{fig:SAM-pose2seg-probl-viskpts}
\end{figure}

\end{document}